\documentclass{isprs} 
\usepackage{subfigure}
\usepackage{subcaption}
\usepackage{setspace}
\usepackage{geometry} 
\usepackage{epstopdf}
\usepackage[labelsep=period]{caption}  
\usepackage[T1]{fontenc}
\usepackage[utf8]{inputenc} 
\usepackage[british]{babel} 
\usepackage[hang]{footmisc}

\usepackage{amsmath, amssymb}

\usepackage{hyperref}
\usepackage{cuted}
\usepackage{capt-of}

\usepackage{graphicx}
\usepackage{tikz}
\usetikzlibrary{calc,arrows.meta,decorations.pathreplacing}
\usepackage{float}

\usepackage{enumitem}

\usepackage[hyphenbreaks]{breakurl}
\usepackage{xurl}

\usepackage{xcolor}
\definecolor{xcol}{RGB}{233,70,21}
\definecolor{ycol}{RGB}{29,233,21}
\definecolor{zcol}{RGB}{21,173,233}
\usetikzlibrary{shadows.blur} 

\definecolor{asg2vmfbg}{RGB}{251,255,229}

\tikzset{
  >={Stealth[length=3mm]},
  arrow/.style   = {-{Stealth}, line width=0.9pt, draw=black},
  train/.style   = {line width=0.9pt, draw=black},
  alabel/.style  = {font=\scriptsize,
                    text opacity=1,
                    rounded corners=0.8pt,
                    inner xsep=2pt, inner ysep=3pt, outer sep=0pt}
}

\usepackage{placeins}
\usepackage{dblfloatfix}

\usepackage[normalem]{ulem}

\usepackage[T1]{fontenc}   
\usepackage{microtype}     

\usepackage{cite}

\usepackage{multirow}
\usepackage{microtype}

\begin{document}

\title{ShinyNeRF: Digitizing Anisotropic Appearance in Neural Radiance Fields}
\date{}

\author{Albert Barreiro\textsuperscript{1,\,2}, Roger Mar\'{i}\textsuperscript{1}, Rafael Redondo\textsuperscript{1}, Gloria Haro\textsuperscript{2}, Carles Bosch\textsuperscript{3}}

\address{
\textsuperscript{1} Eurecat, Centre Tecnològic de Catalunya\\
\textsuperscript{2} Universitat Pompeu Fabra\\
\textsuperscript{3} Universitat de Vic - UCC
}

\vspace{-0.5cm}

\keywords{Neural Radiance Fields, view synthesis, specular reflections, anisotropic material properties, reflectance models}

\abstract{
Recent advances in digitization technologies have transformed the preservation and dissemination of cultural heritage. In this vein, Neural Radiance Fields (NeRF) have emerged as a leading technology for 3D digitization, delivering representations with exceptional realism. However, existing methods struggle to accurately model anisotropic specular surfaces, typically observed, for example, on brushed metals. In this work, we introduce ShinyNeRF, a novel framework capable of handling both isotropic and anisotropic reflections. Our method is capable of jointly estimating surface normals, tangents, specular concentration, and anisotropy magnitudes of an Anisotropic Spherical Gaussian (ASG) distribution, by learning an approximation of the outgoing radiance as an encoded mixture of isotropic von Mises-Fisher (vMF) distributions. Experimental results show that ShinyNeRF not only achieves state-of-the-art performance on digitizing anisotropic specular reflections, but also offers plausible physical interpretations and editing of material properties compared to existing methods.
}

\maketitle

\vspace{-0.5cm}

\section{Introduction}
\label{sec:intro}

Brushed metals, textured fabrics, or treated glass are common anisotropic materials that exhibit complex reflection patterns and are frequently found in cultural heritage collections. Accurate digital reproduction of these materials is essential for the preservation and dissemination of cultural heritage.

Neural rendering techniques, such as Neural Radiance Fields (NeRF) methods~\cite{mildenhall2020nerf} or Gaussian Splatting \cite{kerbl3Dgaussians}, represent the state-of-the-art for digitizing and rendering 3D scenes from image collections. They enable dense, fine-grained 3D representation of objects along with the scene illumination, surpassing classic photogrammetry. These methods rely on consistent appearance across views, performing best in Lambertian scenes with uniform illumination. However, they often struggle with shiny objects, exhibiting strong view-dependent specular reflections.

Specular reflections on anisotropic materials remain a persistent challenge in neural rendering. Most existing approaches rely on non-interpretable learned features to model view-dependent appearance, rather than physical scene properties such as material parameters or ambient illumination. Advanced methods  
incorporate physically grounded reflectance models primarily designed for isotropic behavior and therefore struggle to capture anisotropic reflections, largely due to the absence of reliable surface tangent information.
This problem is also exacerbated by the lack of public datasets containing anisotropic objects with accurate, fully represented ground-truth material properties, which hinders the rigorous evaluation of existing methods. Common digital 3D object collections provide mesh geometry and texture (ShapeNet~\cite{chang2015shapenet}, Objaverse~\cite{deitke2023objaverse}), and sometimes rendered RGB views, depth and normal maps (DISN~\cite{xu2019disn}, Shiny Dataset ~\cite{verbin2022refnerf}) but generally do not include per-point or per-pixel tangent directions.

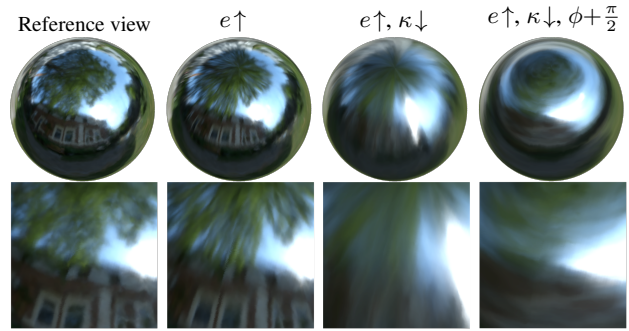
\begin{figure}[t]
\centering
\begin{tikzpicture}
  \node[anchor=south west, inner sep=0] (img)
    {\includegraphics[width=0.99\linewidth]{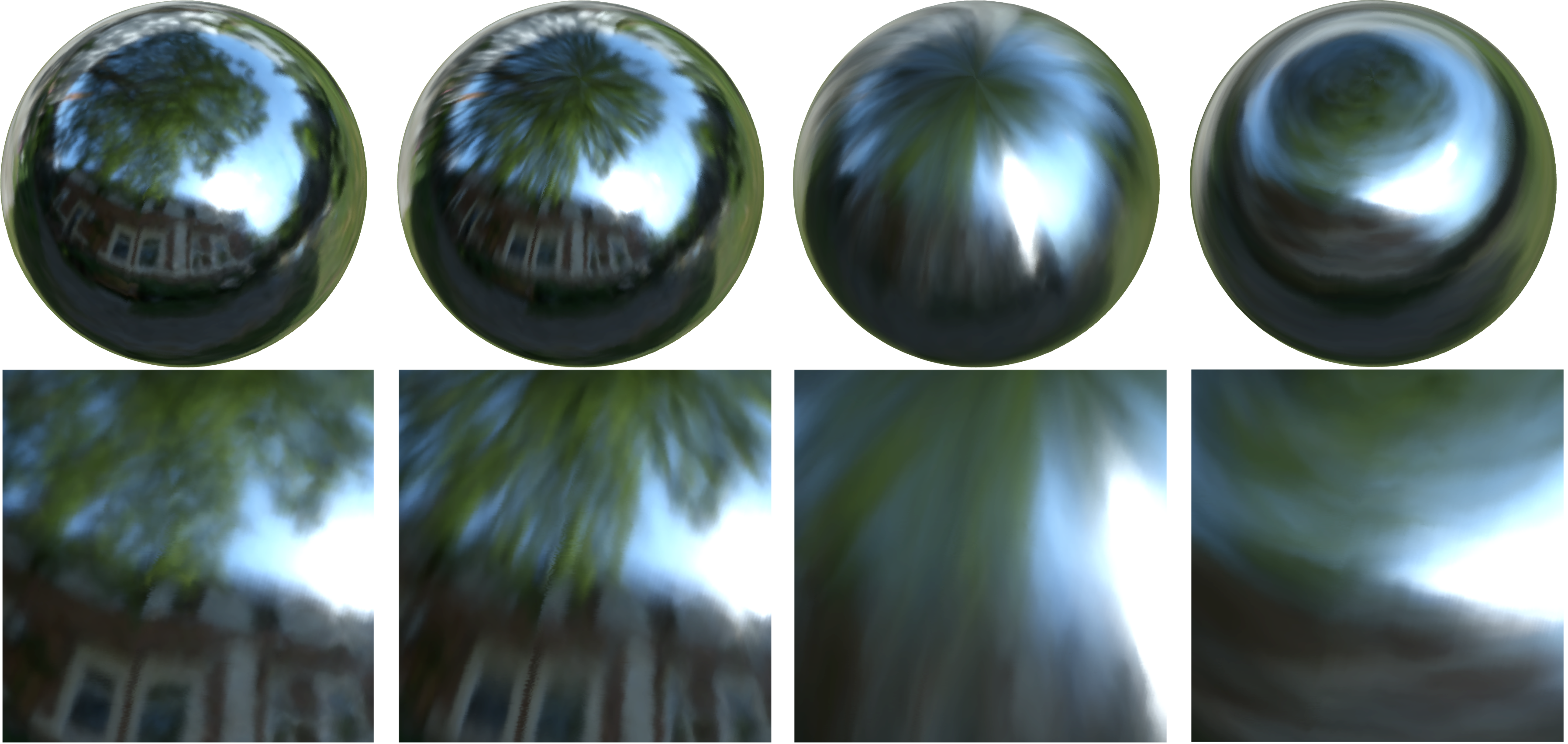}};
  \begin{scope}[x={(img.south east)}, y={(img.north west)}]
  
    \tikzset{col/.style={anchor=south, align=center, inner sep=1pt}}
    \node[col] at (0.12, 1.01)  {\footnotesize Reference view};
    \node[col] at (0.36, 1.01)  {$e \! \uparrow$};
    \node[col] at (0.62, 1.01)  {$e \! \uparrow$, $\kappa \! \downarrow$};
    \node[col] at (0.88, 1.01)  {$e \! \uparrow$, $\kappa \! \downarrow$, $\phi$ \!\!\! $+\frac{\pi}{2}$};
  \end{scope}
\end{tikzpicture}
\caption{ShinyNeRF novel view synthesis varying the learned reflectance parameters. Anisotropy $e$ controls elongation of specular highlights, concentration $\kappa$ reduces the reflection sharpness and the tangent $\phi$ sets the anisotropy orientation.}
\label{fig:teaser}
\end{figure}

This work introduces \textit{ShinyNeRF}, the first NeRF variant to model both isotropic and anisotropic specular reflections within a unified, physically-grounded reflectance framework. While achieving geometry reconstruction and specular rendering comparable to state-of-the-art approaches, ShinyNeRF is the only method offering interpretable material parameters that enable physical understanding and material editing of both isotropic and anisotropic effects, as shown in Figure~\ref{fig:teaser}.

\captionsetup{skip=8pt}

The main contributions of this work are:
\begin{itemize}[itemsep=0pt, topsep=-0pt]
  \item[-] \textbf{Anisotropic reflectance formulation for radiance fields.}
  An anisotropic specular model for NeRF parameterized by concentration, anisotropy, and tangent orientation, with differentiable tangent–bitangent–normal reflection frames.

  \item[-] \textbf{Ground-truth synthetic benchmark with anisotropic material properties.}
  We release two object datasets providing 3D geometry, texture, normals, tangents, and anisotropy parameters for quantitative evaluation of anisotropic view synthesis and estimated material properties.
\end{itemize}

ShinyNeRF code, demos and data are available at \textcolor{blue}{\url{https://multimedia-eurecat.github.io/ShinyNeRF/}}.
\section{Related Work}
\label{sec:related_work}
This section reviews neural scene representations for photorealistic novel view synthesis, focusing on the NeRF formulation, view-dependent appearance modeling, and orthonormal basis construction methods most relevant to our approach for anisotropic neural rendering.

\noindent
\paragraph{NeRF Formulation} 
NeRF~\cite{mildenhall2020nerf} optimizes a neural network encoding the appearance and geometry of a 3D scene by rendering the color observed by the camera rays of an input set of views. The network has two components: (i) a spatial MLP that maps 3D positions $\mathbf{x}$ to a non-negative density $\tau(\mathbf{x})$ and a feature vector $\mathbf{b}(\mathbf{x})$, and (ii) a directional MLP that outputs RGB radiance $\mathbf{c}(\mathbf{x},\hat{\mathbf d})$ given $\mathbf{b}(\mathbf{x})$ and the unit view direction $\hat{\mathbf d}$. For a camera ray $\mathbf{r}(t)=\mathbf{o}+t\,\hat{\mathbf d}$ sampled at depths $\{t_i\}$ with intervals $\Delta t_i=t_{i+1}-t_i$, let $\tau_i$ denote the density at the sample point $\mathbf{r}(t_i)$ and let $\mathbf{c}_i$ denote the RGB color at that point and direction. Volume rendering along the ray yields:
\begin{equation}\label{eq:nerf-discrete}
\begin{aligned}
C(\mathbf o,\hat{\mathbf d}) &= \sum_i w_i\,\mathbf c_i,\quad \\
 \quad  w_i &= e\!^{-\sum_{j<i}\tau_j\,\Delta t_j} \Big(1 - e^{-\tau_i \Delta t_i}\Big) .
\end{aligned}
\end{equation}
where $w_i$ weights the contribution of each point in the camera ray to the rendered color $C(\mathbf o,\hat{\mathbf d})$.

The network parameters are optimized with an $\ell_2$ photometric loss with respect to the ground-truth observed color $C_{\mathrm{gt}}(\mathbf{o},\hat{\mathbf d})$,
\begin{equation}\label{eq:nerf-l2}
\mathcal{L}_{\text{rgb}}=\sum_{(\mathbf{o},\hat{\mathbf d})}\big\|\,C(\mathbf{o},\hat{\mathbf d})-C_{\mathrm{gt}}(\mathbf{o},\hat{\mathbf d})\,\big\|_2^2.
\end{equation}

\paragraph{View-Dependent Appearance}
Approaches to modeling view-dependent effects in neural scene representations can be categorized into three main strategies.

\emph{Multi-lobe encodings} approximate bidirectional reflectance distribution functions (BRDFs) as weighted sums of pre-oriented lobesfrom a shared, predefined bank of directions (or orthonormal axes).
By relying on this fixed lobe set, they remain memory-efficient and fast to evaluate, since only per-point weights or lobe bandwidths are predicted. 
However, the fixed orientations constrain expressiveness, so capturing sharp specular highlights or strong anisotropy typically requires many lobes. 
Because the lobe decomposition is not tied to the local surface frame, view dependence is often absorbed into latent appearance features rather than explicit reflectance parameters, limiting physical interpretability and controlled relighting or material editing. Methods following this approach include NRFF ~\cite{han2023multiscale}, Spec-Gaussian ~\cite{NEURIPS2024_708e0d69}, and AniSDF~\cite{gao2025anisdf}.

\emph{Reflected-ray querying methods} use reconstructed surface geometry to compute reflected rays and query specular appearance along them, rather than regressing an explicit physical BRDF. NeRF-Casting~\cite{verbin2024nerfcasting} traces reflected cones through the scene, aggregating features along each cone and decoding them via a small MLP. SpecNeRF~\cite{ma2023specnerf} instead models specular appearance by encoding reflected ray directions with a learnable Gaussian directional encoding, using roughness to control the Gaussians’ scale, and decoding the resulting embedding to predict specular color.

\emph{Analytic reflectance models}, most notably Ref-NeRF~\cite{verbin2022refnerf} and NeRO~\cite{liu2023nero}, regress parameters of physically motivated reflectance functions at each spatial point. By tying view dependence to analytic reflectance laws, these methods generalize better to novel lighting and viewing conditions.
Ref-NeRF reparameterizes view dependence using reflection directions and structures outgoing radiance into diffuse and specular components together with surface roughness.
This is supported by the isotropic Integrated Directional Encoding (IDE), a closed-form encoding of spherical harmonics integrated under a von Mises-Fisher (vMF) distribution that yields smooth, roughness-conditioned representations of reflected radiance.
NeRO extends this approach to reflective objects by combining a neural SDF surface with an explicit microfacet BRDF. It employs a two-stage optimization: geometry is first recovered using split-sum and IDE-based approximations, then fixed while environment lighting and BRDF parameters are refined through Monte Carlo sampling.

Most of these methods assume isotropic reflectance, where the BRDF $f_r$ depends only on the angles between incoming ($\boldsymbol\omega_i$) and outgoing ($\boldsymbol\omega_o$) light directions relative to the surface normal $\mathbf{n}$, and is invariant to rotations around $\mathbf{n}$. Formally, $f_r(\mathbf{n},\boldsymbol\omega_i,\boldsymbol\omega_o) = f_r(\mathbf{n}, \mathcal{R}_\mathbf{n} \boldsymbol\omega_i, \mathcal{R}_\mathbf{n} \boldsymbol\omega_o)$ for any rotation $\mathcal{R}_\mathbf{n}$ about $\mathbf{n}$. This assumption fails for anisotropic materials such as brushed metals, hair, and woven fabrics, where reflectance depends on both the view angle and the surface tangent orientation.
Modeling such materials requires both a consistent tangent estimation and BRDF parameterization that capture directional anisotropy~\cite{ward1992measuring,burley2012physically}.
Forcing anisotropic effects through isotropic parameterizations leads to geometry deformations that attempt to compensate for missing anisotropic reflectance, rather than achieving geometrically consistent reflectance~\cite{barreiro2025specularity}.

\noindent
\paragraph{ONB from a Single Normal}
\label{subsec:frame-related}
The reflection behavior at anisotropic surface points is naturally tied to the orthonormal basis (ONB) defined by the normal, tangent and bitangent directions. A central challenge of ONB construction methods from a single unit normal is the presence of discontinuities on the unit sphere, which leads to discontinuous tangent parameterizations and noisy gradients near the corresponding singular regions. Classical approaches select an auxiliary direction and complete the basis using cross products~\cite{Hughes01011999}. Faster variants with fewer operations typically introduce branches or special-case handling for numerical stability~\cite{Frisvad01082012}, whereas branch-free formulations of higher algebraic complexity still retain a reduced but unavoidable singular region~\cite{Duff2017Basis}. For neural rendering of anisotropic reflections, ideal ONB constructions should offer explicit control over tangent orientation while remaining smoothly differentiable.

\section{Methodology}

The ShinyNeRF architecture is shown in Figure \ref{fig:architecture}.
Similar to NeRF-based methods~\cite{mildenhall2020nerf},
our method uses a spatial MLP to estimate a dense representation of the diffuse color component $\mathbf{c}_d$ and specular tint $\mathbf{s}$,
a volume density $\tau$, and a bottleneck feature vector $\mathbf{b}$. In addition, ShinyNeRF extends the spatial MLP to estimate the following material parameters: the surface normal $\hat{\mathbf{n}}'$, the specular concentration $\kappa \in [1, \infty)$ as in Ref-NeRF~\cite{verbin2022refnerf}, and crucially for anisotropic reflections, the anisotropy coefficient $e \in [0, 1]$, and the tangent orientation angle $\phi \in [0, \pi]$. These parameters are used to analytically calculate an IDE vector that compactly encodes the outgoing radiance function. The IDE output together with the bottleneck vector $\mathbf{b}$ feed a subsequent directional MLP, responsible for the prediction of the specular color component $\mathbf{c}_s$.
The rendered color $\mathbf{c}$ of each camera ray is finally obtained from the combination of the diffuse and specular components in this manner: $\mathbf{c} = \mathbf{c}_d + \mathbf{s} \odot \mathbf{c}_s$.

\begin{figure}[t]
\centering
\includegraphics[width=\linewidth]{figures/Tref-NeRF_arch.png}
\caption{
ShinyNeRF architecture. The spatial MLP is extended with additional parameters for the anisotropic reflectance model (anisotropy coefficient $e$, tangent orientation $\phi$). The outgoing specular radiance is encoded by a pretrained ASG2vMF network and then forwarded to the directional MLP responsible for rendering the specular color $\mathbf{c}_s$. Yellow blocks represent trainable MLPs, while blue blocks are non-learnable analytic expressions.}
\label{fig:architecture}
\end{figure}

\setlength{\abovedisplayskip}{4pt}
\setlength{\belowdisplayskip}{4pt}

\paragraph{Anisotropic Reflectance Model}
\label{par:anisotropic_model}
To extend the modeling capability of Ref-NeRF, the BRDF is based on the Anisotropic Spherical Gaussian (ASG)~\cite{Xu13sigasia} distribution,
\begin{equation}
ASG(\mathbf{v}; [\mathbf{x}, \mathbf{y}, \mathbf{z}], [\lambda, \mu], c)
= c \cdot \max(\mathbf{v}  \cdot  \mathbf{z} , 0) \cdot 
e^{-\lambda (\mathbf{v} \cdot \mathbf{x})^2 - \mu (\mathbf{v} \cdot \mathbf{y})^2},
\label{eq:asg}
\end{equation}
where $\mathbf{v}$ is the input direction, $\mathbf{z}$, $\mathbf{x}$, $\mathbf{y}$ are the orthonormal \textit{lobe}, \textit{tangent} and \textit{bitangent} axes, respectively; $\lambda$ and $\mu$ are the lobe bandwidths along the $\mathbf{x}$- and $\mathbf{y}$-axes, and $c$ is the lobe amplitude.

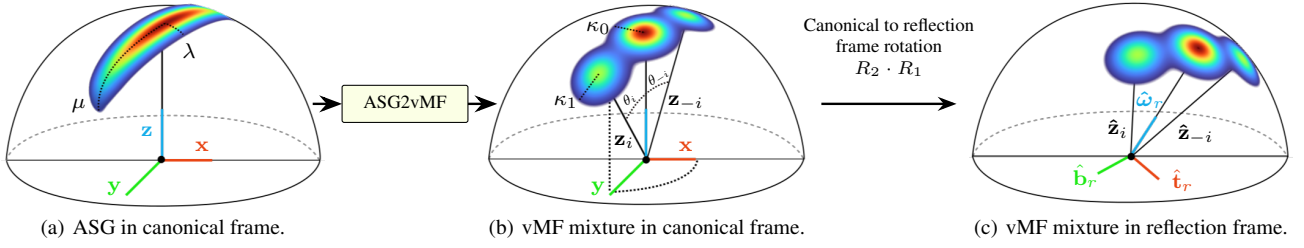
\begin{figure*}[t]
\centering
\subfigure[ASG in canonical frame.]{%
\begin{tikzpicture}[remember picture]
  \node[anchor=south west, inner sep=0] (imgA)
    {\includegraphics[width=.25\textwidth]{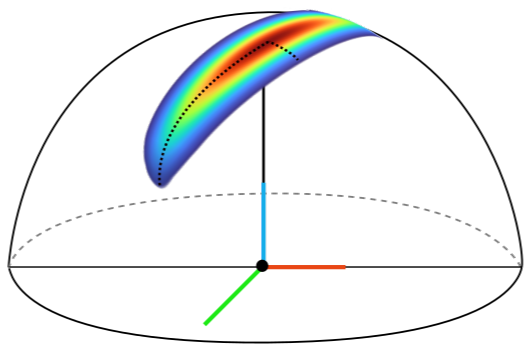}};
  \begin{scope}[x={(imgA.south east)}, y={(imgA.north west)}]
    \path[use as bounding box] (0,0) rectangle (1,1);
    \node[font=\footnotesize]  at (0.58,0.75) {$\lambda$};
    \node[font=\footnotesize]  at (0.24,0.49) {$\mu$};
    \node[font=\footnotesize, text=xcol] at (0.62,0.3) {$\mathbf{x}$};
    \node[font=\footnotesize, text=ycol] at (0.35,0.10) {$\mathbf{y}$};
    \node[font=\footnotesize, text=zcol] at (0.45,0.37) {$\mathbf{z}$};
  \end{scope}
\end{tikzpicture}}%
\hfill
\subfigure[vMF mixture in canonical frame.]{%
\begin{tikzpicture}[remember picture]
  \node[anchor=south west, inner sep=0] (imgB)
    {\includegraphics[width=.25\textwidth]{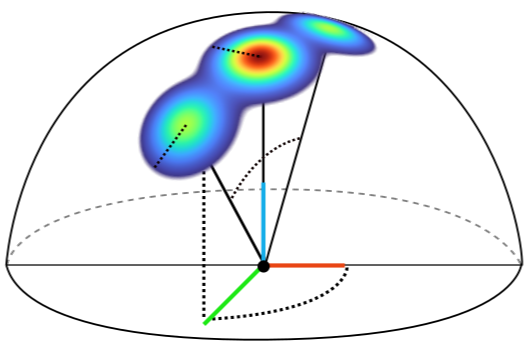}};
  \begin{scope}[x={(imgB.south east)}, y={(imgB.north west)}]
    \path[use as bounding box] (0,0) rectangle (1,1);
    \node[anchor=south] at (0.35,0.79) {\footnotesize$\kappa_0$};
    \node[anchor=south] at (0.25,0.44) {\footnotesize$\kappa_1$};
    \node[anchor=south, rotate=33, scale=0.6, transform shape] at (0.464,0.48) {$\theta_{i}$}; 
    \node[anchor=south, rotate=33, scale=0.6, transform shape] at (0.555,0.58) {$\theta_{-i}$}; 

    \node[font=\footnotesize, text=xcol] at (0.62,0.3) {$\mathbf{x}$};
    \node[font=\footnotesize, text=ycol] at (0.35,0.10) {$\mathbf{y}$};   
    \node[anchor=south] at (0.43,0.25) 
    {\footnotesize$\mathbf{z}_i$};
    \node[anchor=south] at (0.62,0.43)
    {\footnotesize$\mathbf{z}_{-i}$};
  \end{scope}
\end{tikzpicture}}%
\hfill
\subfigure[vMF mixture in reflection frame.]{%
\begin{tikzpicture}[remember picture]
  \node[anchor=south west, inner sep=0] (imgC)
    {\includegraphics[width=.25\textwidth]{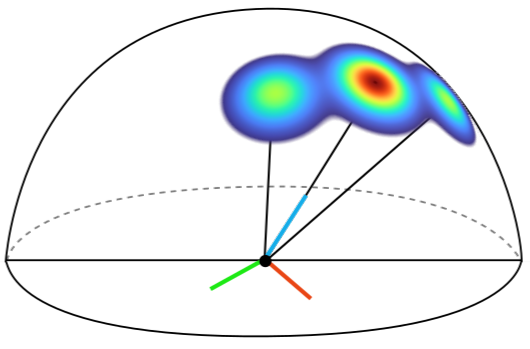}};
  \begin{scope}[x={(imgC.south east)}, y={(imgC.north west)}]
    \path[use as bounding box] (0,0) rectangle (1,1);
    \node[font=\footnotesize, text=xcol] at (0.66,0.14) {$\hat{\mathbf{t}}_r$};
    \node[font=\footnotesize, text=ycol] at (0.36,0.16) {$\hat{\mathbf{b}}_r$};
    \node[font=\footnotesize, text=zcol] at (0.56,0.52) {$\boldsymbol{\hat\omega}_r$};
        \node[anchor=south] at (0.46,0.3) 
    {\footnotesize$\mathbf{\hat{z}}_i$};
    \node[anchor=south] at (0.7,0.27)
    {\footnotesize$\mathbf{\hat{z}}_{-i}$};
  \end{scope}
\end{tikzpicture}}%

\begin{tikzpicture}[remember picture,overlay]
  \coordinate (startAB) at ($(imgA.east)+(-0.5em,0)$);
  \coordinate (endAB)   at ($(imgB.west)+(0.5em,0)$);

  \path (startAB) -- (endAB) coordinate[pos=0.5] (midAB);

  \node[alabel, fill=asg2vmfbg, draw=black] (asg2vmf) at (midAB) {{\begin{tabular}{c}
         ASG2vMF 
    \end{tabular}}};

  \draw[train,->] (startAB) -- (asg2vmf.west);
  \draw[train,->] (asg2vmf.east) -- (endAB);

  \draw[arrow]
    ($(imgB.east)+(0.5em,0)$) --
    node[alabel, above=6pt] {\begin{tabular}{c}
         Canonical to reflection \\ frame rotation \\$R_2 \cdot R_1$
    \end{tabular}}
    ($(imgC.west)+(-0.5em,0)$);
\end{tikzpicture}

\caption{ShinyNeRF anisotropic reflection model: the pretrained ASG2vMF network approximates the ASG ($\lambda$, $\mu$) in a canonical frame as a symmetric mixture of vMF functions, each defined by a concentration $\kappa_i$, an elevation $\theta_i$, and a weight $\alpha_i$ (not depicted). 
Vectors in red-green-blue denote orthonormal basis.
The mean direction $\mathbf{z}_i$ of the vMFs is
finally rotated according to the direction of reflection $\boldsymbol{\hat\omega}_r$. }
\label{fig:ide-construction}
\end{figure*}

We derive the axial bandwidths from the predicted anisotropy $e$ and concentration $\kappa$ coefficients as follows:
\begin{equation}
\mu = \tfrac{1}{2}\,\kappa\,(1-e),
\qquad
\lambda = \tfrac{1}{2}\,\kappa\,(1+e),
\end{equation}
where the isotropic case reduces to $e = 0$ and $\mu = \lambda$. 
The scalar $e \in [0,1]$ acts as an anisotropy control: larger values increase the difference between $\lambda$ and $\mu$, producing more elongated specular lobes. Without loss of generality, we set $\lambda \geq \mu$, so the tangent axis $\mathbf{x}$ corresponds to the minor bandwidth and the bitangent axis $\mathbf{y}$ to the major bandwidth of the lobe.

\paragraph{ASG Approximation via vMF Mixture}
Building upon the IDE in Ref-NeRF, this work represents the function of outgoing radiance by approximating the ASG distribution as a mixture of $N$ isotropic von Mises-Fisher (vMF) distributions. Representing elongated anisotropic shapes with mixtures of isotropic lobes or kernels is a well-established technique in classic anisotropic reflectance modeling and texture filtering~\cite{kautz2000approximation,mccormack1999feline,wang2009all}.

First, consider the following vMF formulation
\begin{equation}
\begin{aligned}
\mathrm{vMF}(\mathbf{v}; \mathbf{z}, \kappa)=
\mathcal{N}(\kappa)\exp\Bigl(\kappa\, \mathbf{z} \cdot \mathbf{v} \Bigr),
\end{aligned}
\label{eq:vmfl}
\end{equation}
where $\mathbf{z}$ and $\kappa$ are the vMF mean direction and concentration, respectively, and $\mathcal{N}(\kappa)$ is a normalization constant on $\mathbb{S}^2$.

Then, we define the vMFs mixture approximating an ASG as:
\begin{equation}
\begin{aligned}
\hat{p}_{\text{ASG}}(\mathbf{v}; \{\alpha_i,\mathbf{z}_i, \kappa_i\})=
\sum_{i=-L}^{L} \alpha_i\,
\mathrm{vMF}(\mathbf{v}; \mathbf{z}_i, \kappa_i),
\end{aligned}
\label{eq:vmf-mix}
\end{equation}
where $\alpha_i$ are normalized mixture weights for $N=2L+1$ vMFs, and $\mathbf{z}_i=R_\mathbf{x}(\theta_i)\,\mathbf{z}$ is the orientation of the $i$-th vMF with respect to the $\mathbf{z}$ axis, rotated by an angle $\theta_i$ around the $\mathbf{x}$-axis. 
To make the approximation independent of lobe orientation, both the $N$ vMFs and the ASG are expressed in the canonical frame, $[\mathbf{x}, \mathbf{y}, \mathbf{z}]$ in Equation~\eqref{eq:asg}, aligning the principal direction of the ASG lobe with the $+\mathbf{z}$ axis (north pole). The $N$ vMF components of the mixture are then placed along the $\mathbf{y}$-axis of major bandwidth, where the ASG exhibits greater elongation. Figure~\ref{fig:ide-construction}(a--b) illustrates a toy example with $N=3$ vMF.

To reduce the number of variables needed in Equation~\eqref{eq:vmf-mix} and enforce symmetry, we center a vMF towards $\mathbf{z}_0=\mathbf{z}$ with $\theta_0=0$ and concentration $\kappa_0$, and the rest of vMFs are arranged as two symmetrical lateral sets of size $L$, satisfying that $\theta_{-i}=-\theta_i$, $\kappa_{-i}=\kappa_i$, and $\alpha_{-i}=\alpha_i$. Thus, the probability distribution $\hat{p}_{\text{ASG}}$ can be parametrized by a vector $\hat{\mathbf{q}}$ of size $3L + 2$,
\begin{equation}
      \hat{\mathbf{q}} = [\{\theta_i, \kappa_i, \alpha_i\}_{i=1}^{L}, \kappa_0,\alpha_0].
    \label{eq:vmf_mixture_params}
\end{equation}

A lightweight auxiliary MLP, referred to as ASG2vMF, is employed to estimate the parameters $\hat{\mathbf{q}}$ in Equation~\eqref{eq:vmf_mixture_params} defining the vMF mixture modeling from the ASG with bandwidths $\lambda$ and $\mu$.
For numerical stability, the ASG2vMF receives a vector of log-bandwidths as input: $[\log\lambda, \log\mu, \log(\mu/\lambda)]$. The log-ratio $\log(\mu/\lambda)$ characterizes the degree of anisotropy, with $\log(\mu/\lambda)=0$ corresponding to the isotropic case where $\mu=\lambda$. Each log-component passes through a positional encoding with 10 frequency bands for capturing high-frequency details.

The ASG2vMF network is optimized by minimizing the Kullback–Leibler~\cite{Kullback1951OnIA} and Jensen–Shannon~\cite{JS_distribution} divergences:
\begin{equation}
\mathcal{L}_\mathrm{asg} = \text{KL}(p_{\text{ASG}} \| \hat{p}_{\text{ASG}}) + \beta \cdot \text{JS}(p_{\text{ASG}}, \hat{p}_{\text{ASG}}),
\label{eq:pretrain-loss}
\end{equation}
where $p_{\text{ASG}}(\mathbf{v};\lambda,\mu)$ is the normalized ASG target probability density function defined over the upper hemisphere $\mathbf{v}\cdot\mathbf{z}\ge 0$.
The weight $\beta=0.3$ controls the JS contribution, which acts as a symmetric and bounded regularizer that discourages degenerate mixtures, whereas the KL term drives a mode-covering fit to the target $p_{\text{ASG}}$.
In practice, both distributions are evaluated on a fixed grid over the main lobe hemisphere and normalized by the sum of their values weighted by each cell solid angle, ensuring the integral sums one and it is directly comparable.
During training of the ASG2vMF, the ASG bandwidth range is set to $\lambda, \mu \in [10^{-2}, 600]$ with $\lambda \geq \mu$, and arbitrary combinations of $(\lambda, \mu)$ are sampled. During ShinyNeRF optimization, the ASG2vMF weights are frozen.

\begin{figure}
  \centering
  \includegraphics[width=1\linewidth]{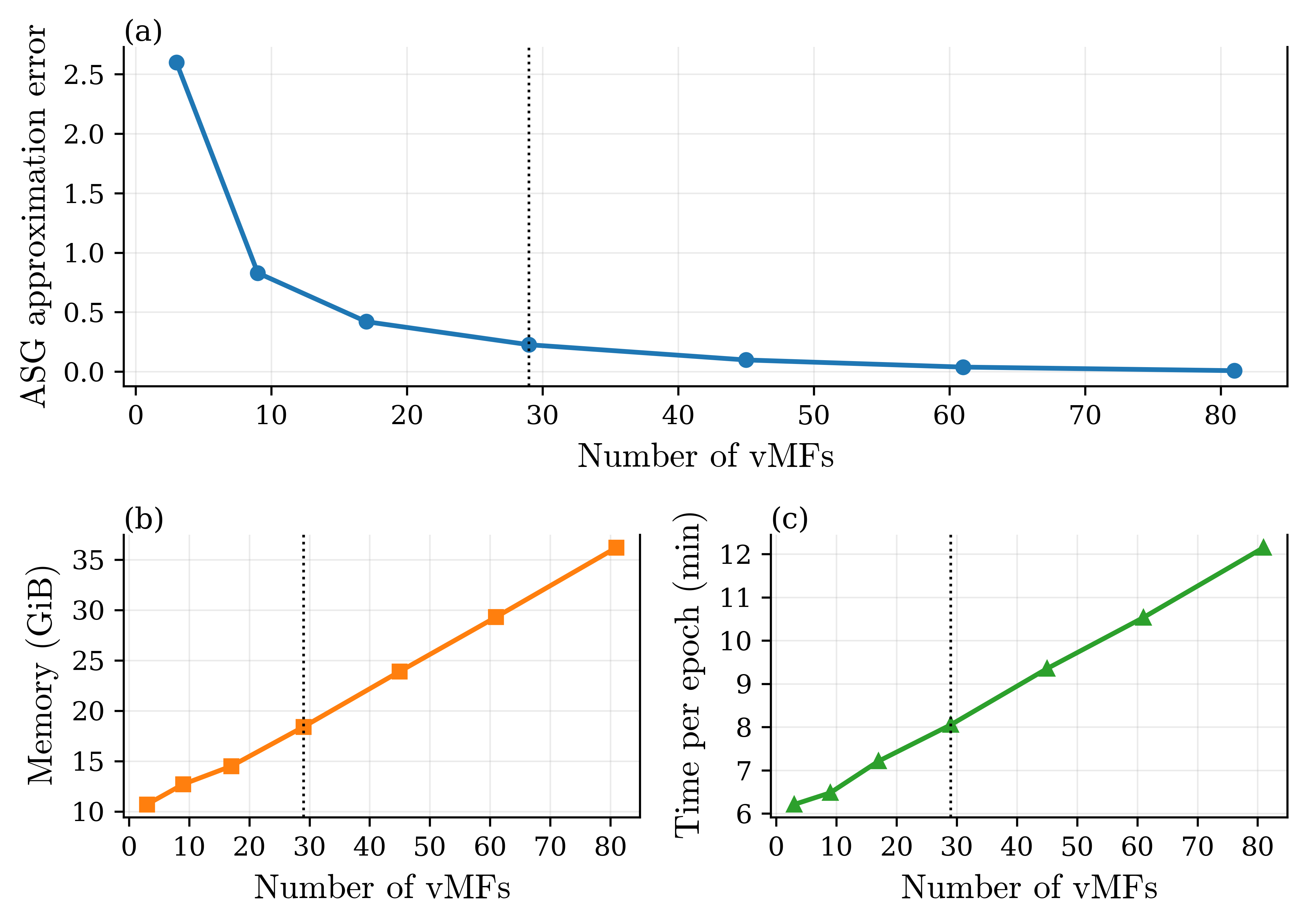}
  \caption{(a) ASG approximation error, (b)~memory cost and (c)~time per epoch for ShinyNeRF with varying number of vMF components $N$. The vertical dotted line marks the chosen number $N = 29$. Reported values use a training batch size of 3000 rays.} 
  \label{fig:vMF_vs_loss}
\end{figure}

\begin{figure}[t]
\centering
\begin{tikzpicture}
  \node[anchor=south west, inner sep=0] (img)
     {\includegraphics[width=0.95\linewidth]{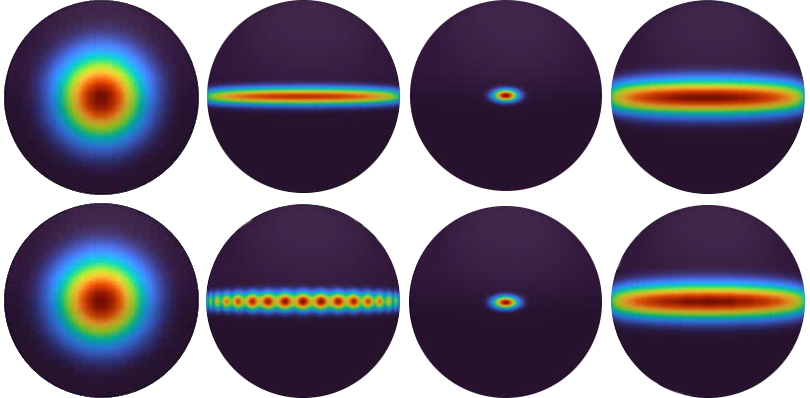}};

  \node[anchor=north, inner sep=0] (barimg)
     at ([yshift=-2mm]img.south) 
     {%
       \includegraphics[
         width=0.02\linewidth,    
         height=0.9\linewidth,   
         keepaspectratio=false,
         angle=270,
         origin=c,
         clip
       ]{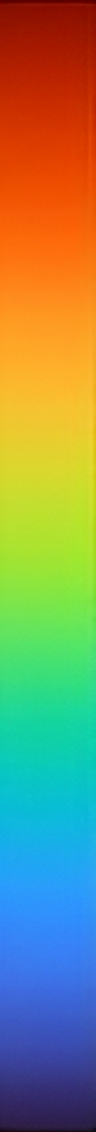}%
     };

  \node[anchor=north, font=\scriptsize] at ([yshift=36mm]barimg.south west) {$0$}; 
  \node[anchor=north, font=\scriptsize] at ([yshift=36mm]barimg.south east) {$1$};
    
  \begin{scope}[x={(img.south east)}, y={(img.north west)}]
    \tikzset{col/.style={anchor=south, align=center, inner sep=1pt, font=\small}}
    \foreach \i/\lam/\mui in {1/10/10, 2/270/0.01, 3/600/133, 4/66/0.01}{
      \pgfmathsetmacro\x{(2*\i-1)/8}
      \node[col] at (\x, 1.00) {$\lambda=\lam$\\$\mu=\mui$};
    }
    \node[rotate=90, font=\scriptsize] at (-0.05,0.76) {ASG};
    \node[rotate=90, font=\scriptsize] at (-0.05,0.24) {vMF mixture};

    \draw[->, thick] (-0.015,-0.01) -- (0.23,-0.01)
      node[anchor=west, font=\scriptsize] {$y$};
    \draw[->, thick] (-0.015,-0.01) -- (-0.015,0.41)
      node[anchor=south, font=\scriptsize] {$x$};

  \end{scope}
\end{tikzpicture}
\vspace{-35mm}

\caption{ASG approximation with $N=29$ vMFs.}
\label{fig:asg2vmf}
\end{figure}
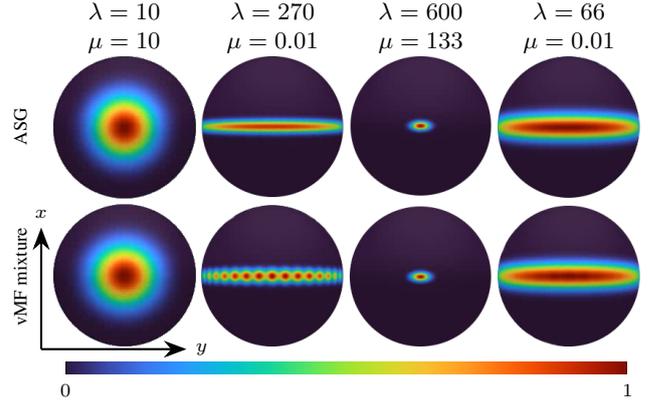

The number of vMFs involves a trade-off between precision and computational efficiency, as shown in Figure~\ref{fig:vMF_vs_loss}. The ASG approximation error decreases rapidly with the number of components, being $N=29$ a reasonable middle-point.
Figure~\ref{fig:asg2vmf} shows a visual comparison of the ASG approximation with a vMF mixture for different anisotropic bandwidth values.

\paragraph{Anisotropic Orientation}

The ASG-based BRDF, approximated by the mixture of isotropic vMFs, is reoriented according to the orthonormal basis composed by the surface normal $\hat{\mathbf{n}}'$, surface tangent $\hat{\mathbf{t}}$, and bitangent $\hat{\mathbf{b}}$. This ONB is constructed from the predicted $\hat{\mathbf{n}}'=(n_x,n_y,n_z)$ and the tangent orientation $\phi$. The tangent is initialized to $\hat{\mathbf{t}}_0$, a vector belonging to the plane $z=0$ and orthogonal to $\hat{\mathbf{n}}'$,  and the bitangent must be orthogonal to $\hat{\mathbf{n}}'$ and $\hat{\mathbf{t}}_0$. It is calculated as follows: 
\begin{equation}
\hat{\mathbf{t}}_0 = \frac{(-n_y, n_x, 0) \, \,}{\|(-n_y, n_x, 0)\|_2}, \quad \hat{\mathbf{b}}_0 = \hat{\mathbf{n}}'\times \hat{\mathbf{t}}_0.
\end{equation}
To steer the vector basis towards the final anisotropic orientation $\phi$, the following rotation is applied in the normal plane:
\begin{equation}
\hat{\mathbf{t}} = \cos\phi\,\hat{\mathbf{t}}_0 + \sin\phi\,\hat{\mathbf{b}}_0, \quad
\hat{\mathbf{b}} = -\sin\phi\,\hat{\mathbf{t}}_0 + \cos\phi\,\hat{\mathbf{b}}_0.
\end{equation}
Note that this ad-hoc controllable ONB construction is not continuous at the poles, if $\hat{\mathbf{n}}'=(0,0,\pm1)$. In practice, these special cases are handled by a small tolerance epsilon.

\paragraph{IDE Construction}

The anisotropic IDE used in ShinyNeRF is derived from the $N = 2L + 1$ vMF distributions that approximate the ASG function of outgoing radiance as 
\begin{equation}
\begin{aligned}
\mathrm{IDE}_{\text{ASG}}=
\sum_{i=-L}^L \alpha_i\;
\mathrm{IDE}_{\text{vMF}}(\hat{\mathbf{z}}_i, \kappa_i),
\end{aligned}
\label{eq:ide-mix}
\end{equation}
where $\mathrm{IDE}_{\text{vMF}}$ is the isotropic IDE closed-form expression introduced in Ref-NeRF~\cite{verbin2022refnerf} after the vMF distribution. Each vMF is characterized by $\alpha_i$, $\hat{\mathbf{z}}_i$ and $\kappa_i$ (weight, mean direction and concentration, respectively), obtained after the ASG approximation via vMF mixture.

In Equation~\eqref{eq:ide-mix}, the vMF mixture (or the ASG lobe, equivalently) is reoriented from the canonical frame towards the direction of reflection $\hat{\boldsymbol{\omega}}_r$. Following Ref-NeRF, $\hat{\mathbf{d}}$ is the unit ray direction from camera to surface, and the reflection direction is computed as $\hat{\boldsymbol{\omega}}_r = 2 (-\hat{\mathbf{d}} \cdot \hat{\mathbf{n}}') \hat{\mathbf{n}}' + \hat{\mathbf{d}}$ using the predicted normal $\hat{\mathbf{n}}'$. Thus, the mean direction $\hat{\mathbf{z}}_i$ of each vMF in the reflection frame is a rotated version of $\mathbf{z}_i$ from Equation~\eqref{eq:vmf-mix}.

As shown in Figure~\ref{fig:ide-construction}(c), the rotation that expresses $\mathrm{IDE}_{\text{ASG}}$ in the reflection frame amounts to two successive rotations: $R_2 R_1 $. The first rotation matrix $R_1$ aligns the canonical ASG axes with the surface frame, i.e., $[\hat{\mathbf{t}}\;\hat{\mathbf{b}}\;\hat{\mathbf{n}}']=R_1[\mathbf{x}\;\mathbf{y}\;\mathbf{z}]$.
Then, following the Rodrigues' formulation, the second rotation $R_2$ transforms the surface frame to align the normal with the reflection direction $\hat{\boldsymbol{\omega}}_r$, i.e., $[\hat{\mathbf{t}}_r\;\hat{\mathbf{b}}_r\;\hat{\boldsymbol{\omega}}_r]=R_2\,[\hat{\mathbf{t}}\;\hat{\mathbf{b}}\;\hat{\mathbf{n}}']$.

Note that processing each $\mathrm{IDE}_{\text{vMF}}$ involved in Equation~\eqref{eq:ide-mix} independently would require multiple forward passes of the directional MLP, significantly increasing computational cost. Alternative multi-lobe approaches concatenate all lobe encodings~\cite{han2023multiscale,NEURIPS2024_708e0d69,gao2025anisdf}, increasing the encoding size and consequently the computational overhead. In contrast, the weighted summation in Equation~\eqref{eq:ide-mix} follows directly from the linearity of the IDE integral, requiring only one forward pass per ray while maintaining a compact encoding dimensionality.

\paragraph{Loss Function}

ShinyNeRF is trained using the photometric reconstruction loss $\mathcal{L}_\text{rgb}$ from Equation~\eqref{eq:nerf-l2} combined with additional geometric regularization loss terms. The complete loss function is
\begin{equation}
\begin{aligned}
\mathcal{L}_\text{ShinyNeRF}
&= \mathcal{L}_{\text{rgb}}
+ \beta_\text{pred} \mathcal{L}_{\text{pred}}
+ \beta_\text{grad} \mathcal{L}_{\text{grad}}
\\
&\quad
+ \beta_\text{orient} \mathcal{L}_{\text{orient}}
+ \beta_\text{dist} \mathcal{L}_{\text{dist}}
+ \beta_\text{prop} \mathcal{L}_{\text{prop}},
\label{eq:loss_complete}
\end{aligned}
\end{equation}
where the hyperparameters  $\beta_\text{pred}=3 \times 10^{-3}$, $\beta_\text{grad}=3 \times 10^{-4}$, $\beta_\text{orient}=0.01$, $\beta_\text{dist}=3 \times 10^{-3}$ and  $\beta_\text{prop}= 3 \times 10^{-4}$ are empirically chosen weights for the auxiliary terms.

\textit{Asymmetric Normal Loss} -- This term is adopted from NeRF-Casting~\cite{verbin2024nerfcasting} to enforce consistency between the predicted normals $\hat{\mathbf{n}}'$ and the geometric normals $\hat{\mathbf{n}}$ (defined as the normalized negative gradient of the volume density $\tau$), penalizing deviations in both directions:
\begin{equation}
\begin{aligned}
\mathcal{L}_\text{pred}
&=
\mathcal{L}_\text{cos}\bigl(\text{sg}(w), \text{sg}(\hat{\mathbf{n}}), \hat{\mathbf{n}}'\bigr)
\\
\mathcal{L}_\text{grad}
&=
\mathcal{L}_\text{cos}\bigl(w, \hat{\mathbf{n}}, \text{sg}(\hat{\mathbf{n}}')\bigr),
\end{aligned}
\label{eq:asymetric_normal_loss}
\end{equation}
where sg(·) denotes stop-gradient and
\begin{equation}
\mathcal{L}_\text{cos}(w, \hat{\mathbf{n}}, \hat{\mathbf{n}}')
=
\sum_i
w_i
\bigl(1 - \hat{\mathbf{n}}_i \cdot \hat{\mathbf{n}}'_i\bigr)
\end{equation}
penalizes different angle orientation between the normalized $\hat{\mathbf{n}}_i$ and $\hat{\mathbf{n}}'_i$ along each ray according to the volume rendering weights $w_i$. $\mathcal{L}_\text{cos}$ is equivalent, up to a constant factor, to the squared $\ell_2$ used in NeRF-Casting for unit-norm normals.

$\mathcal{L}_\text{pred}$ in Equation~\eqref{eq:asymetric_normal_loss} aligns $\hat{\mathbf{n}}'$ with the geometry encoded by $\hat{\mathbf{n}}$ and $w$ derived from the density $\tau$, while $\mathcal{L}_{\text{grad}}$ reciprocally updates $\hat{\mathbf{n}}$ and $w$ toward $\hat{\mathbf{n}}'$. The loss is asymmetric because of the different weights, $\beta_\text{pred}$ and $\beta_\text{grad}$, used in Equation~\eqref{eq:loss_complete}.

\textit{Normal Orientation Loss} -- Following Ref-NeRF~\cite{verbin2022refnerf}, $\mathcal{L}_\text{orient}$ is adopted to penalize back-facing normals (i.e., predicted normals pointing away from the camera):
\begin{equation}
\mathcal{L}_\text{orient}
=
\sum_i
w_i \,\max\bigl(0,\hat{\mathbf{n}}'_i \cdot \hat{\mathbf{d}}\bigr)^2.
\end{equation}
This term encourages visible samples along each camera ray (those with large $w_i$) to satisfy $\hat{\mathbf{n}}'_i \cdot \hat{\mathbf{d}} < 0$, enforcing normals that point towards the camera (opposite to the view direction $\hat{\mathbf{d}}$).

\textit{Distortion Loss} -- Following mip-NeRF 360~\cite{barron2022mipnerf360}, $\mathcal{L}_\mathrm{dist}$ is used to reduce floater artifacts and fragmented geometry results that can occur in NeRF approaches. For each sample position $t_i$ with weight $w_i$ along a ray, it minimizes
\begin{equation}
\mathcal{L}_\mathrm{dist}
=
\sum_{i,j}
w_i w_j
\left|
\frac{t_i + t_{i+1}}{2}
-
\frac{t_j + t_{j+1}}{2}
\right|
+
\frac{1}{3}
\sum_i
w_i^2 (t_{i+1} - t_i).
\end{equation}
The first term in $\mathcal{L}_\mathrm{dist}$ penalizes broadly distributed density or separated density peaks along the ray, and the second term promotes narrow, concentrated intervals where weights $w_i$ are large.

\textit{Proposal Loss} -- Following Zip-NeRF~\cite{barron2023zipnerf}, a lightweight auxiliary MLP is used to estimate a coarse distribution of volume density for efficient sampling, restricting ShinyNeRF evaluations to regions with non-negligible density. Let $(t,w)$ and $(\hat{t},\hat{w})$ denote the sample positions and volume rendering weights of NeRF and the proposal MLP along a ray, respectively. The fine-scale samples $(t,w)$ are drawn from the coarse proposal distribution $(\hat{t},\hat{w})$, with weights $\hat{w}$ predicted by the proposal MLP, which is trained with
\begin{equation}
\mathcal{L}_\text{prop}
= 
\sum_i
\frac{1}{\hat{w}_i}
\Bigl[
\max\bigl(0,\,
\text{sg}(w_i^{\hat{t}}) - \hat{w}_i
\bigr)
\Bigr]^2.
\end{equation}
$\mathcal{L}_\text{prop}$ encourages a smooth, anti-aliased proposal distribution by penalizing proposal weights $\hat{w}_i$ that fall below corresponding smoothed NeRF weights $w_i^{\hat{t}}$, obtained by blurring and resampling $(t,w)$ at positions $\hat{t}$.

\section{Experiments}
\label{sec:experiments}
This section presents the evaluation of ShinyNeRF and the proposed anisotropic reflectance model. The conducted experiments validate three key claims: (1) the proposed tangent estimation enables stable learning of anisotropic materials, (2) the method accurately reconstructs directional specular highlights on both simple and complex geometries, and (3) the approach generalizes across diverse lighting conditions and material parameters.

\paragraph{Implementation Details}
We used 8 hidden layers of width 256 and ReLU activations both in the spatial MLP and directional MLP of ShinyNeRF (Figure~\ref{fig:architecture}). The input of the spatial MLP follows an integrated positional encoding of 3D samples $\mathbf{x}$ as in~\cite{barron2022mipnerf360}, while the directional MLP takes as input the 32-dimensional bottleneck feature vector, the 72-dimensional anisotropic IDE encoding (degree 5) and the \mbox{1-dimensional} dot product $\hat{\mathbf{d}} \cdot \hat{\mathbf{n}}'$. The auxiliary proposal MLP (4 hidden layers with 256 units and ReLU activations) is trained jointly with ShinyNeRF, while the auxiliary ASG2vMF MLP (3 hidden layers with 128 units and LeakyReLU activations) is pre-trained offline and kept frozen during ShinyNeRF optimization.

ShinyNeRF was trained on two NVIDIA RTX 3090 GPUs with a batch size of 6000, requiring approximately 30 GB of GPU memory. Training employs the Adam optimizer with $\beta_1=0.9$ (momentum) and $\beta_2=0.999$ (second moment decay for adaptive learning rates). The learning rate is initialized at $5.0 \times 10^{-4}$ and decays exponentially to a final value of $5.0 \times 10^{-6}$ after a warm-up period of 2500 steps with sine scheduling. Gradients are clipped to a maximum norm of $1.0 \times 10^{-3}$ to ensure training stability. With this configuration, training an object takes around 20 hours, reducing training time by 80\% compared to Ref-NeRF.

\paragraph{Datasets}
Evaluation is performed on three synthetic datasets that target different aspects of anisotropic appearance modeling. All scenes contain 68 training views and 33 test views.

\noindent\textit{Anisotropic Synthetic Dataset (ASD)} was introduced in Spec-Gaussian~\cite{NEURIPS2024_708e0d69} and comprises a set of objects with pronounced specular reflections under direct point lighting without environment maps. Anisotropic material parameters are roughly constant for each object. Since the original dataset included only RGB views, we re-rendered the objects in Blender to generate normal and tangent maps as ground-truth geometry. All views and maps have a resolution of $270{\times}270$ pixels. This dataset is used to evaluate the method on varying geometries under controlled illumination and homogeneous reflectance behavior.\footnote{The ASD scenes used for evaluation are: \textit{teapot}, \textit{ashtray}, \textit{jupyter}, \textit{dishes}.}

\noindent\textit{Anisotropic Spheres (ASPH)} is a novel dataset created and publicly released for this work. It consists of four spheres rendered under different environment maps, with systematic variations in anisotropic material parameters (anisotropy strength and specular concentration). All views and maps are rendered at a resolution of $270{\times}270$ pixels. This dataset is used to evaluate the performance of the method on simple geometries under complex, varying environment illumination and reflectance behavior.

\noindent\textit{Cultural Heritage Anisotropic Objects (CHAO)} is a second dataset introduced in this work. It comprises a Moroccan barad teapot and a German medieval helmet, both exhibiting strong anisotropic specularities. All views and maps are rendered at a resolution of $512{\times}512$ pixels. This dataset is used to assess the performance of the method on real cultural heritage objects under complex geometry and environment lighting. The 3D objects were obtained from the Sketchfab catalog.\footnote{\textcolor{blue}{\url{https://sketchfab.com}}.}

Using synthetic objects provides access to ground-truth anisotropic material properties defined in Blender. However, Blender’s anisotropy strength and specular concentration are not directly comparable to the equivalent anisotropic parameters learned by ShinyNeRF ($e$ and $\kappa$, respectively), due to fundamental differences in BRDF parameterization and reflectance models. Consequently, we rely on qualitative analysis to assess whether the predicted anisotropic parameters reproduce the patterns observed in the corresponding ground-truth properties, and perform quantitative evaluation only on the output RGB renderings and the estimated normal and tangent maps.

\paragraph{Evaluation Metrics}
Image quality is evaluated using standard metrics in neural rendering: PSNR, SSIM~\cite{wang2004ssim}, and LPIPS~\cite{zhang2018lpips}. Geometric consistency of the reconstructed objects is assessed using the mean angular error (MAE$^\circ$) in degrees of the estimated normal and tangent vectors.

\begin{table}[t]
\centering
\setlength{\tabcolsep}{4pt}\renewcommand{\arraystretch}{1.4}
\resizebox{\columnwidth}{!}{%
\begin{tabular}{c|c|ccc|ccc}
\hline
\rule{0pt}{2.4ex}
 & \multirow{2}{*}{Method} 
 & \multirow{2}{*}{PSNR $\uparrow$} 
 & \multirow{2}{*}{SSIM $\uparrow$} 
 & \multirow{2}{*}{LPIPS $\downarrow$} 
 & \multicolumn{3}{c}{MAE$^\circ$ $\downarrow$} \\
 &  &  &  &  & $\hat{\mathbf{n}}$ & $\hat{\mathbf{n}}'$ & $\hat{\mathbf{t}}$ \\
\hline
\multirow{4}{*}{\rotatebox[origin=c]{90}{ASD}} 
  & Ref-NeRF   &29.51 & 0.851 & 0.259 & 42.83 & 23.83 & - \\
  & Spec Gaussian   & 30.66 & \textbf{0.941} & \textbf{0.118} & - & 52.75 & -  \\
  & AniSDF & 32.23 & 0.930 & 0.185 & - & \textbf{16.23} & - \\
  & ShinyNeRF  &  \textbf{32.66} & 0.938 & 0.176 & \textbf{29.01} & 23.61 & \textbf{48.14}   \\
\hline
\multirow{4}{*}{\rotatebox[origin=c]{90}{ASPH}} 
  & Ref-NeRF   &  29.93 & 0.929 & 0.144 & 33.57 & 2.16 & - \\
 & Spec Gaussian & 29.15 & 0.914 & 0.123 & - & 56.53 & -  \\
  & AniSDF & 28.21 & 0.867 & 0.224 & - & \textbf{1.11} & -  \\
  & ShinyNeRF  & \textbf{36.42} & \textbf{0.971} & \textbf{0.102} & \textbf{8.86} & 1.33 & \textbf{1.36} \\
\hline
\multirow{4}{*}{\rotatebox[origin=c]{90}{CHAO}} 
  & Ref-NeRF   &  23.72 & 0.799 & 0.257 & 52.10 & 35.58 & - \\
 & Spec Gaussian & 23.78 & \textbf{0.854} & \textbf{0.174} & - & 51.88 & -  \\
  & AniSDF & \textbf{26.01} & 0.850 & 0.206 & - & \textbf{7.75} & -  \\
  & ShinyNeRF  & 25.31 & 0.847 & 0.207 & \textbf{25.89} & 11.46 & \textbf{40.08} \\
\hline
\multirow{4}{*}{\rotatebox[origin=c]{90}{Mean}} 
  & Ref-NeRF   & 27.72  & 0.860 & 0.220 & 42.84 & 20.52 & - \\
 & Spec Gaussian & 27.86 & 0.903 & \textbf{0.138} & - & 53.72 & -  \\
  & AniSDF & 28.82 & 0.882 & 0.205 & - & \textbf{8.36} & -  \\
  & ShinyNeRF  & \textbf{31.46} & \textbf{0.919} & 0.162 & \textbf{21.25} & 12.13 & \textbf{29.86} \\
\end{tabular}
}
\caption{Quantitative evaluation on the ASD, ASPH, and CHAO datasets based on PSNR/SSIM/LPIPS for RGB renderings and MAE$^\circ$ for geometric density-gradient normals $\hat{\mathbf{n}}$, predicted normals $\hat{\mathbf{n}}'$, and tangents $\hat{\mathbf{t}}$ (only available in ShinyNeRF).}
\label{tab:quantitative_results}
\end{table}

\subsection{Results}

\begin{figure}[ht!]
  \centering
  \begin{tabular}{@{\hspace{0cm}}c@{\hspace{0.1cm}}c@{\hspace{0.1cm}}c@{\hspace{0.1cm}}c@{\hspace{0.1cm}}c@{\hspace{0.1cm}}c}
  & \scriptsize GT & \scriptsize ShinyNeRF & \scriptsize AniSDF & \scriptsize Spec-Gaussian & \scriptsize Ref-NeRF \\

  
   \multirow{2}{*}{%
    \raisebox{0.5\height}{
      \rotatebox[origin=c]{90}{\scriptsize RGB}%
    }} &
  \includegraphics[width=0.18\columnwidth,keepaspectratio,clip,trim=28 80 20 50]{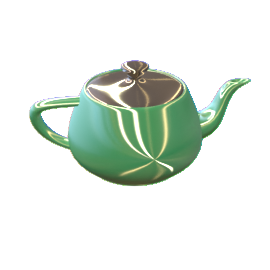} &
  \includegraphics[width=0.18\columnwidth,keepaspectratio,clip,trim=28 80 20 50]{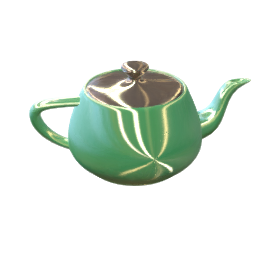} &
  \includegraphics[width=0.18\columnwidth,keepaspectratio,clip,trim=28 80 20 50]{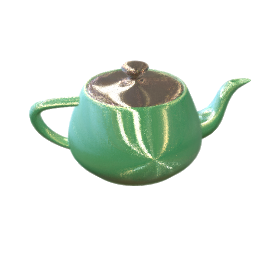} &
  \includegraphics[width=0.18\columnwidth,keepaspectratio,clip,trim=28 80 20 50]{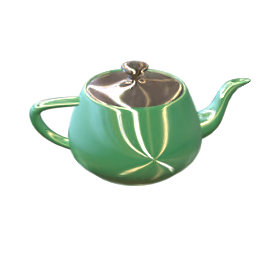} &
  \includegraphics[width=0.18\columnwidth,keepaspectratio,clip,trim=28 80 20 50]{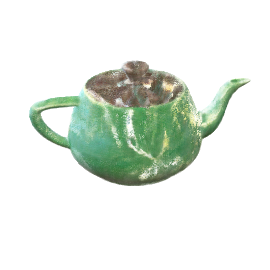} \\
  &
  \includegraphics[width=0.18\columnwidth,keepaspectratio,clip,trim=100 100 90 90]{figures/qualitative/teapot/gt_teapot_rgb.png} &
  \includegraphics[width=0.18\columnwidth,keepaspectratio,clip,trim=100 100 90 90]{figures/qualitative/teapot/shinynerf_teapot_rgb.png} &
  \includegraphics[width=0.18\columnwidth,keepaspectratio,clip,trim=100 100 90 90]{figures/qualitative/teapot/anisdf_teapot_rgb.png} &
  \includegraphics[width=0.18\columnwidth,keepaspectratio,clip,trim=100 100 90 90]{figures/qualitative/teapot/specgaussian_teapot_rgb.png} &
  \includegraphics[width=0.18\columnwidth,keepaspectratio,clip,trim=100 100 90 90]{figures/qualitative/teapot/refnerf_teapot_rgb.png} \\
   \multirow{2}{*}{%
    \raisebox{0.5\height}{
      \rotatebox[origin=c]{90}{\scriptsize Normal $\hat{\mathbf{n}}'$}%
    }} &
  \includegraphics[width=0.18\columnwidth,keepaspectratio,clip,trim=28 80 20 50]{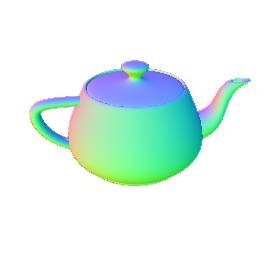} &
  \includegraphics[width=0.18\columnwidth,keepaspectratio,clip,trim=28 80 20 50]{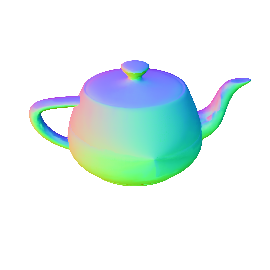} &
  \includegraphics[width=0.18\columnwidth,keepaspectratio,clip,trim=28 80 20 50]{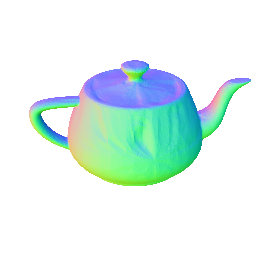} &
  \includegraphics[width=0.18\columnwidth,keepaspectratio,clip,trim=28 80 20 50]{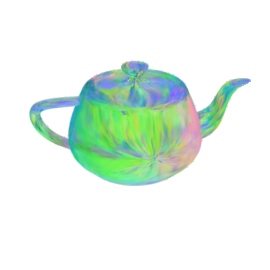} &
  \includegraphics[width=0.18\columnwidth,keepaspectratio,clip,trim=28 80 20 50]{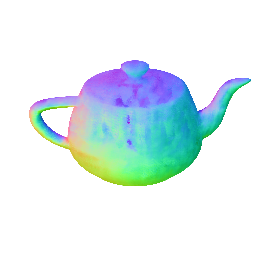} \\
  &
  \includegraphics[width=0.18\columnwidth,keepaspectratio,clip,trim=100 100 90 90]{figures/qualitative/teapot/gt_teapot_normal.png} &
  \includegraphics[width=0.18\columnwidth,keepaspectratio,clip,trim=100 100 90 90]{figures/qualitative/teapot/shinynerf_teapot_normal.png} &
  \includegraphics[width=0.18\columnwidth,keepaspectratio,clip,trim=100 100 90 90]{figures/qualitative/teapot/anisdf_teapot_normal.png} &
  \includegraphics[width=0.18\columnwidth,keepaspectratio,clip,trim=100 100 90 90]{figures/qualitative/teapot/specgaussian_teapot_normal.png} &
  \includegraphics[width=0.18\columnwidth,keepaspectratio,clip,trim=100 100 90 90]{figures/qualitative/teapot/refnerf_teapot_normal.png} \\ \hline \\[-0.8em]

  
   \multirow{2}{*}{%
    \raisebox{0.5\height}{
      \rotatebox[origin=c]{90}{\scriptsize RGB}%
    }} &
  \includegraphics[width=0.18\columnwidth,keepaspectratio]{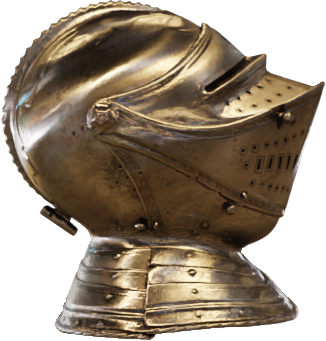} &
  \includegraphics[width=0.18\columnwidth,keepaspectratio]{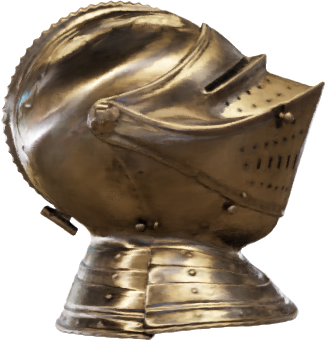} &
  \includegraphics[width=0.18\columnwidth,keepaspectratio]{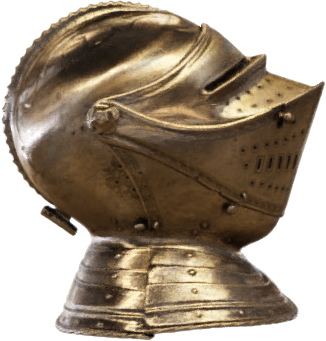}&
  \includegraphics[width=0.18\columnwidth,keepaspectratio]{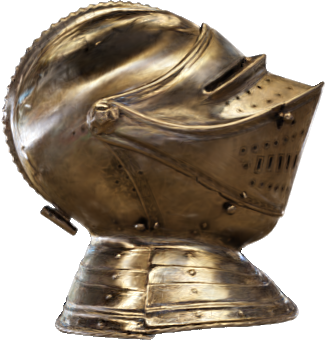} &
  \includegraphics[width=0.18\columnwidth,keepaspectratio]{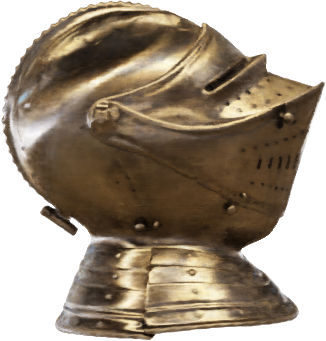} \\
  &
  \includegraphics[width=0.18\columnwidth,keepaspectratio,clip,trim=45 115 90 35]{figures/qualitative/helmet/gt_helmet_rgb.png} &
  \includegraphics[width=0.18\columnwidth,keepaspectratio,clip,trim=45 115 90 35]{figures/qualitative/helmet/shinynerf_helmet_rgb.png} &
  \includegraphics[width=0.18\columnwidth,keepaspectratio,clip,trim=45 115 90 35]{figures/qualitative/helmet/anisdf_helmet_rgb.png}&
  \includegraphics[width=0.18\columnwidth,keepaspectratio,clip,trim=45 115 90 35]{figures/qualitative/helmet/specgaussian_helmet_rgb.png} &
  \includegraphics[width=0.18\columnwidth,keepaspectratio,clip,trim=45 115 90 35]{figures/qualitative/helmet/refnerf_helmet_rgb.png} \\
   \multirow{2}{*}{%
    \raisebox{0.5\height}{
      \rotatebox[origin=c]{90}{\scriptsize Normal $\hat{\mathbf{n}}'$}%
    }} &
  \includegraphics[width=0.18\columnwidth,keepaspectratio]{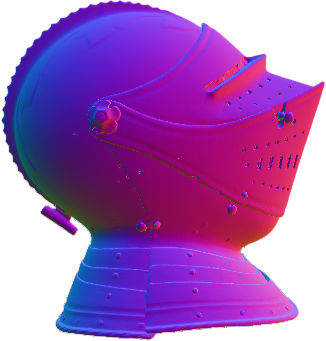} &
  \includegraphics[width=0.18\columnwidth,keepaspectratio]{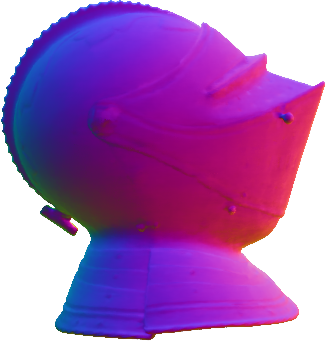} &
  \includegraphics[width=0.18\columnwidth,keepaspectratio]{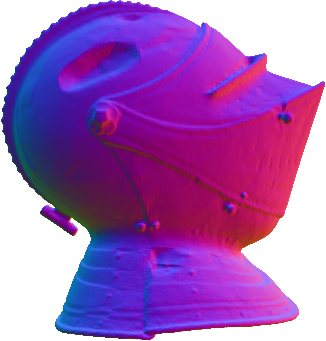}&
  \includegraphics[width=0.18\columnwidth,keepaspectratio]{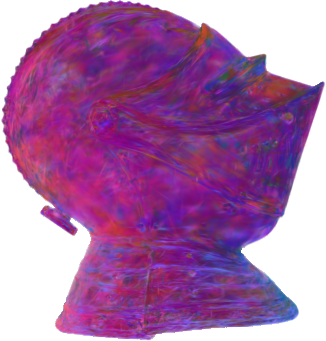} &
  \includegraphics[width=0.18\columnwidth,keepaspectratio]{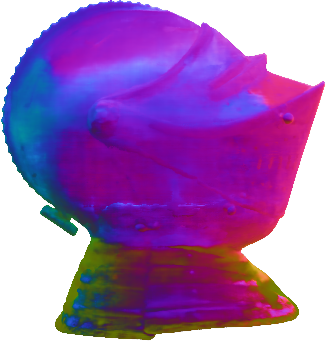} \\
  &
  \includegraphics[width=0.18\columnwidth,keepaspectratio,clip,trim=45 115 90 35]{figures/qualitative/helmet/gt_helmet_normal.png} &
  \includegraphics[width=0.18\columnwidth,keepaspectratio,clip,trim=45 115 90 35]{figures/qualitative/helmet/shinynerf_helmet_normal.png} &
  \includegraphics[width=0.18\columnwidth,keepaspectratio,clip,trim=45 115 90 35]{figures/qualitative/helmet/anisdf_helmet_normal.png}&
  \includegraphics[width=0.18\columnwidth,keepaspectratio,clip,trim=45 115 90 35]{figures/qualitative/helmet/specgaussian_helmet_normal.png} &
  \includegraphics[width=0.18\columnwidth,keepaspectratio,clip,trim=45 115 90 35]{figures/qualitative/helmet/refnerf_helmet_normal.png} \\ \hline \\[-0.8em]

  
   \multirow{2}{*}{%
    \raisebox{0.5\height}{
      \rotatebox[origin=c]{90}{\scriptsize RGB}%
    }} &
  \includegraphics[width=0.18\columnwidth,keepaspectratio,clip,trim=75 102 95 65]{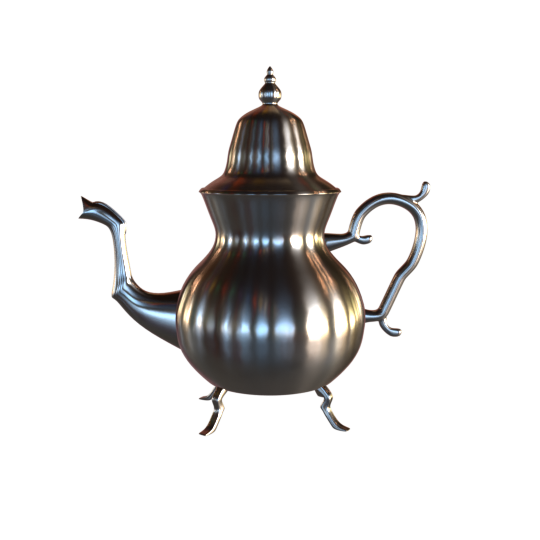} &
  \includegraphics[width=0.18\columnwidth,keepaspectratio,clip,trim=75 102 95 65]{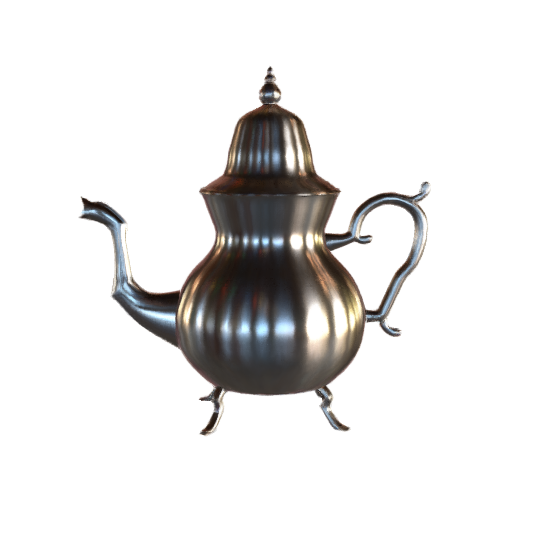} &
  \includegraphics[width=0.18\columnwidth,keepaspectratio,clip,trim=75 102 95 65]{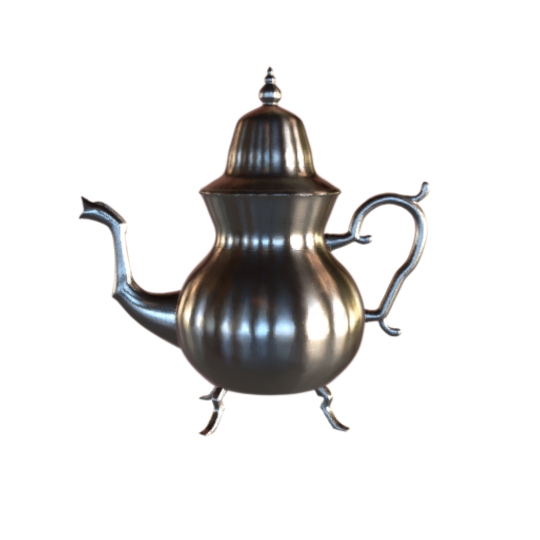} &
  \includegraphics[width=0.18\columnwidth,keepaspectratio,clip,trim=75 102 95 65]{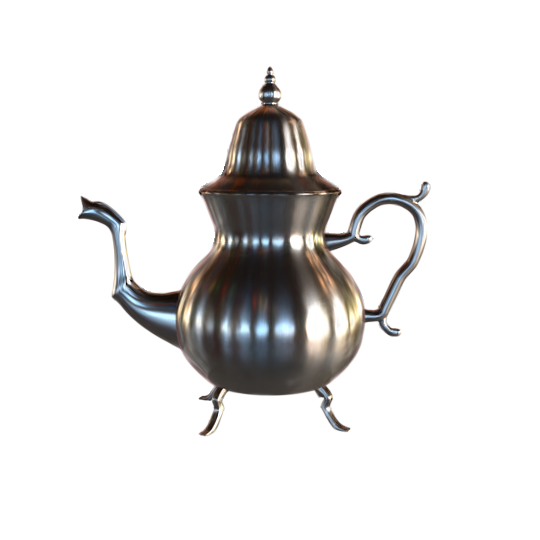} &
  \includegraphics[width=0.18\columnwidth,keepaspectratio,clip,trim=75 102 95 65]{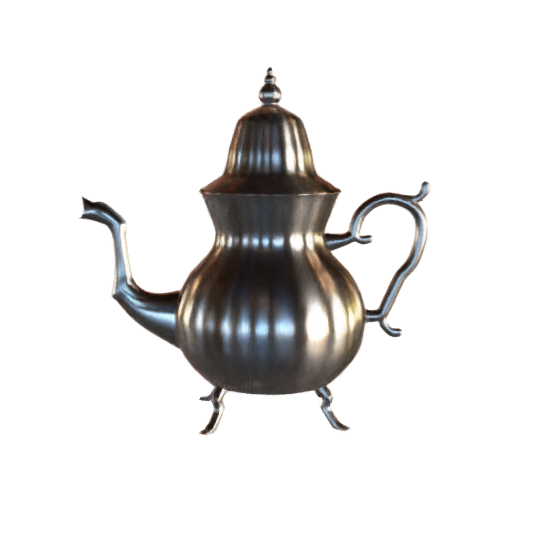} \\
  &
  \includegraphics[width=0.18\columnwidth,keepaspectratio,clip,trim=240 170 200 270]{figures/qualitative/barad/gt_rgb.png} &
  \includegraphics[width=0.18\columnwidth,keepaspectratio,clip,trim=240 170 200 270]{figures/qualitative/barad/shinynerf_rgb.png} &
  \includegraphics[width=0.18\columnwidth,keepaspectratio,clip,trim=240 170 200 270]{figures/qualitative/barad/anisdf_rgb.png} &
  \includegraphics[width=0.18\columnwidth,keepaspectratio,clip,trim=240 170 200 270]{figures/qualitative/barad/specgaussian_rgb.png} &
  \includegraphics[width=0.18\columnwidth,keepaspectratio,clip,trim=240 170 200 270]{figures/qualitative/barad/refnerf_rgb.png}\\
   \multirow{2}{*}{%
    \raisebox{0.5\height}{
      \rotatebox[origin=c]{90}{\scriptsize Normal $\hat{\mathbf{n}}'$}%
    }} &
  \includegraphics[width=0.18\columnwidth,keepaspectratio,clip,trim=75 102 95 65]{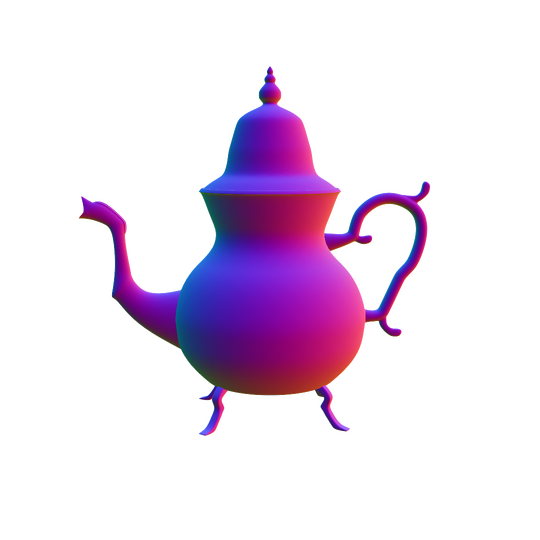} &
  \includegraphics[width=0.18\columnwidth,keepaspectratio,clip,trim=75 102 95 65]{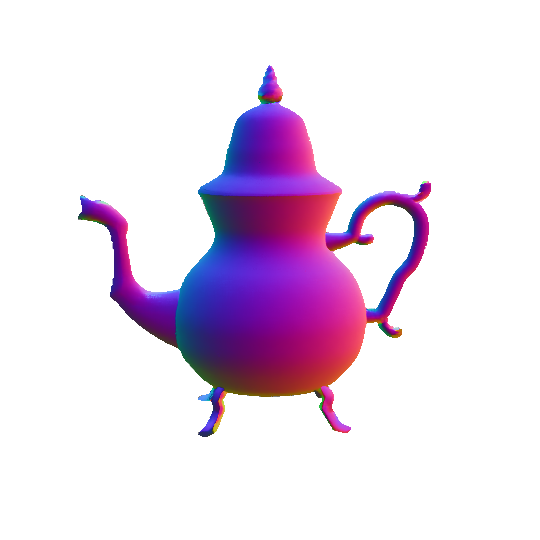} &
  \includegraphics[width=0.18\columnwidth,keepaspectratio,clip,trim=75 102 95 65]{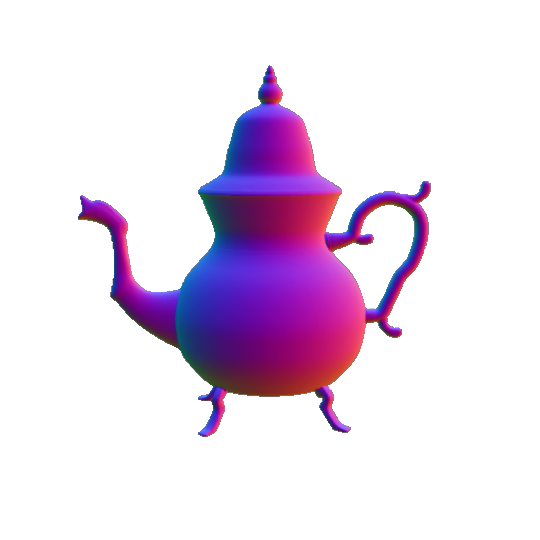}&
  \includegraphics[width=0.18\columnwidth,keepaspectratio,clip,trim=75 102 95 65]{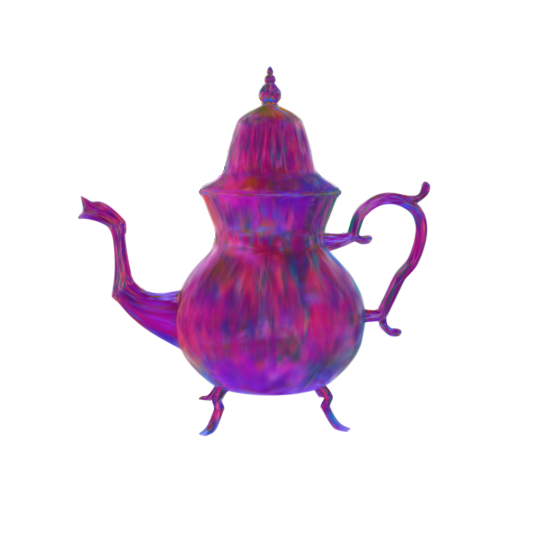} &
  \includegraphics[width=0.18\columnwidth,keepaspectratio,clip,trim=75 102 95 65]{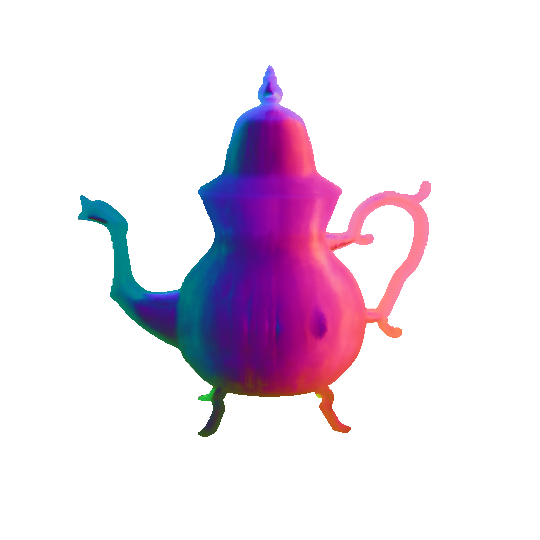} \\
  &
  \includegraphics[width=0.18\columnwidth,keepaspectratio,clip,trim=240 170 200 270]{figures/qualitative/barad/gt_normal2.png} &
  \includegraphics[width=0.18\columnwidth,keepaspectratio,clip,trim=240 170 200 270]{figures/qualitative/barad/shinynerf_normal.png} &
  \includegraphics[width=0.18\columnwidth,keepaspectratio,clip,trim=240 170 200 270]{figures/qualitative/barad/anisdf_normal.png}&
  \includegraphics[width=0.18\columnwidth,keepaspectratio,clip,trim=240 170 200 270]{figures/qualitative/barad/specgaussian_normal.png} &
  \includegraphics[width=0.18\columnwidth,keepaspectratio,clip,trim=240 170 200 270]{figures/qualitative/barad/refnerf_normal.png}

\end{tabular}
  \caption{Qualitative comparison of RGB renderings and predicted normals $\hat{\mathbf{n}}'$. Top to bottom: \textit{teapot} (ASD dataset), \textit{helmet} and \textit{barad teapot} (CHAO dataset).
  }
  \label{fig:teapot_comparison}
\end{figure}

Table~\ref{tab:quantitative_results} reports quantitative results on the test sets of the proposed datasets, comparing ShinyNeRF to reference methods: Ref-NeRF~\cite{verbin2022refnerf}, Spec-Gaussian~\cite{NEURIPS2024_708e0d69}, and AniSDF~\cite{gao2025anisdf}. ShinyNeRF attains the highest PSNR in average, indicating the lowest pixel-wise reconstruction error and closest agreement with the photometric training objective. However, improvements in SSIM and LPIPS are less pronounced; on ASD, Spec-Gaussian achieves the best LPIPS by a noticeable margin. We hypothesize that subtle view-dependent effects and high-frequency content can make perceptual metrics such as LPIPS less tightly correlated with the global rendering quality, leading to less clear trends in their values. Visual inspection of the RGB renderings reveals no significant differences between ShinyNeRF and Spec-Gaussian, as shown in Figure~\ref{fig:teapot_comparison}.

ShinyNeRF also achieves state-of-the-art geometry in both predicted and geometric normals, $\hat{\mathbf{n}}'$ and $\hat{\mathbf{n}}$, respectively. Geometric density-gradient normals $\hat{\mathbf{n}}$ are reported only for NeRF-based methods using a volume density field, i.e., Ref-NeRF and ShinyNeRF in Table~\ref{tab:quantitative_results}. AniSDF achieves the lowest MAE$^\circ$ for predicted normals $\hat{\mathbf{n}}'$, reflecting its generally sharper geometry compared to ShinyNeRF. However, this level of sharpness can also amplify geometric artifacts in the presence of specular reflections that exceed AniSDF’s representational capacity, such as the large cavity it produces on the helmet in Figure~\ref{fig:teapot_comparison}. Overall, we find that ShinyNeRF achieves the best trade-off between artifact-free reconstructions and global geometric accuracy. Methods that ignore anisotropy, such as Ref-NeRF, tend to explain anisotropic reflection effects via incorrect geometry, as reflected in the higher $\hat{\mathbf{n}}$ MAE$^\circ$ in Table~\ref{tab:quantitative_results}.

\begin{figure}[ht!]
  \centering
  \begin{tabular}{@{\hspace{0cm}}c@{\hspace{0.1cm}}c@{\hspace{0.1cm}}c@{\hspace{0.1cm}}c@{\hspace{0.1cm}}c}
  & \scriptsize RGB rendering & \scriptsize Tangent $\hat{\mathbf{t}}$ & \scriptsize Anisotropy $e$ & \scriptsize Concentration $\kappa$ \\

  
  \raisebox{3\height}{\rotatebox[origin=c]{90}{\scriptsize GT}} &
  \includegraphics[width=0.22\columnwidth,keepaspectratio,clip,trim=27 27 27 27]{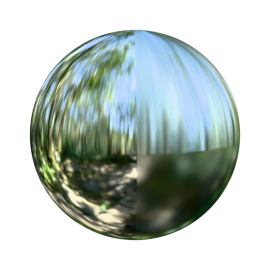} &
  \includegraphics[width=0.22\columnwidth,keepaspectratio,clip,trim=27 27 27 27]{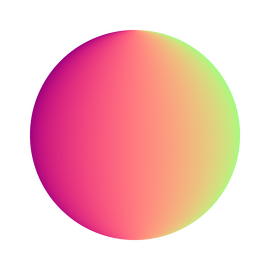} &
  \includegraphics[width=0.22\columnwidth,keepaspectratio,clip,trim=27 27 27 27]{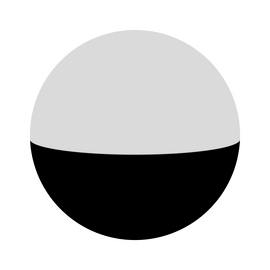} &
  \includegraphics[width=0.22\columnwidth,keepaspectratio,clip,trim=27 27 27 27]{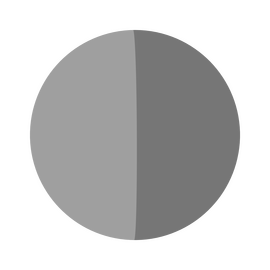} \\
  \raisebox{1.5\height}{\rotatebox[origin=c]{90}{\scriptsize Predicted}} &
  \includegraphics[width=0.22\columnwidth,keepaspectratio,clip,trim=27 27 27 27]{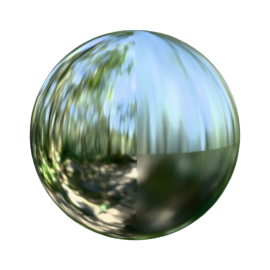} &
  \includegraphics[width=0.22\columnwidth,keepaspectratio,clip,trim=27 27 27 27]{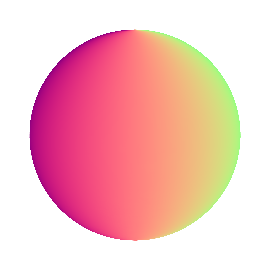} &
  \includegraphics[width=0.22\columnwidth,keepaspectratio,clip,trim=27 27 27 27]{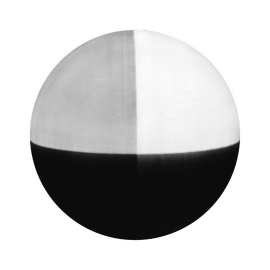} &
  \includegraphics[width=0.22\columnwidth,keepaspectratio,clip,trim=27 27 27 27]{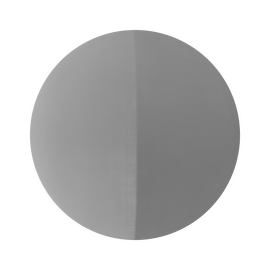} \\ \hline \\[-0.7em]

  
  \raisebox{3\height}{\rotatebox[origin=c]{90}{\scriptsize GT}} &
  \includegraphics[width=0.22\columnwidth,keepaspectratio,clip,trim=27 27 27 27]{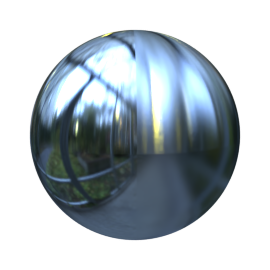} &
  \includegraphics[width=0.22\columnwidth,keepaspectratio,clip,trim=27 27 27 27]{figures/qualitative2/gt_t.png} &
  \includegraphics[width=0.22\columnwidth,keepaspectratio,clip,trim=27 27 27 27]{figures/qualitative2/gt_ecc.png} &
  \includegraphics[width=0.22\columnwidth,keepaspectratio,clip,trim=27 27 27 27]{figures/qualitative2/gt_conc.png} \\
  \raisebox{1.5\height}{\rotatebox[origin=c]{90}{\scriptsize Predicted}} &
  \includegraphics[width=0.22\columnwidth,keepaspectratio,clip,trim=27 27 27 27]{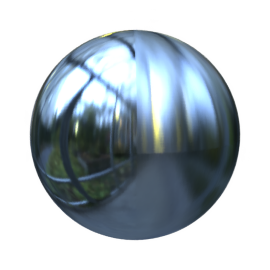} &
  \includegraphics[width=0.22\columnwidth,keepaspectratio,clip,trim=27 27 27 27]{figures/qualitative2/pred_t.png} &
  \includegraphics[width=0.22\columnwidth,keepaspectratio,clip,trim=27 27 27 27]{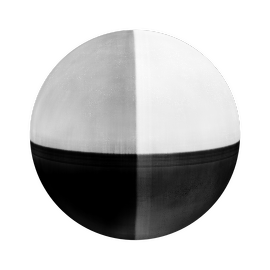}  &
  \includegraphics[width=0.22\columnwidth,keepaspectratio,clip,trim=27 27 27 27]{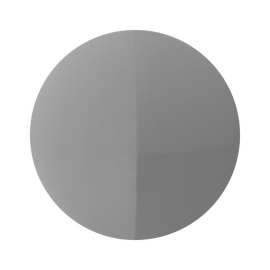} \\ \hline \\[-0.7em]

  
  \raisebox{3\height}{\rotatebox[origin=c]{90}{\scriptsize GT}} &
  \includegraphics[width=0.22\columnwidth,keepaspectratio,clip,trim=27 27 27 27]{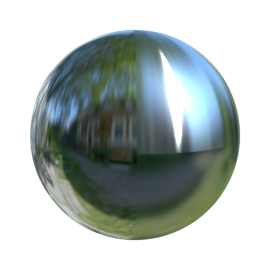} &
  \includegraphics[width=0.22\columnwidth,keepaspectratio,clip,trim=27 27 27 27]{figures/qualitative2/gt_t.png} &
  \includegraphics[width=0.22\columnwidth,keepaspectratio,clip,trim=27 27 27 27]{figures/qualitative2/gt_ecc.png} &
  \includegraphics[width=0.22\columnwidth,keepaspectratio,clip,trim=27 27 27 27]{figures/qualitative2/gt_conc.png} \\
  \raisebox{1.5\height}{\rotatebox[origin=c]{90}{\scriptsize Predicted}} &
  \includegraphics[width=0.22\columnwidth,keepaspectratio,clip,trim=27 27 27 27]{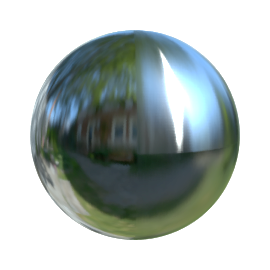} &
  \includegraphics[width=0.22\columnwidth,keepaspectratio,clip,trim=27 27 27 27]{figures/qualitative2/pred_t.png} &
  \includegraphics[width=0.22\columnwidth,keepaspectratio,clip,trim=27 27 27 27]{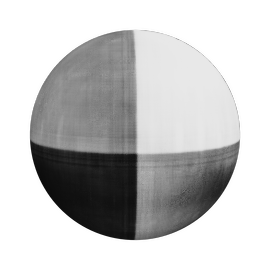}  &
  \includegraphics[width=0.22\columnwidth,keepaspectratio,clip,trim=27 27 27 27]{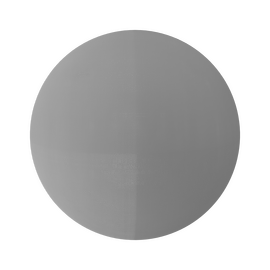} \\ \hline \\[-0.7em]

  
  \raisebox{3\height}{\rotatebox[origin=c]{90}{\scriptsize GT}} &
  \includegraphics[width=0.22\columnwidth,keepaspectratio,clip,trim=27 27 27 27]{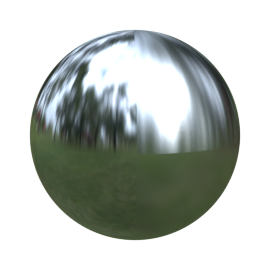} &
  \includegraphics[width=0.22\columnwidth,keepaspectratio,clip,trim=27 27 27 27]{figures/qualitative2/gt_t.png} &
  \includegraphics[width=0.22\columnwidth,keepaspectratio,clip,trim=27 27 27 27]{figures/qualitative2/gt_ecc.png} &
  \includegraphics[width=0.22\columnwidth,keepaspectratio,clip,trim=27 27 27 27]{figures/qualitative2/gt_conc.png} \\
  \raisebox{1.5\height}{\rotatebox[origin=c]{90}{\scriptsize Predicted}} &
  \includegraphics[width=0.22\columnwidth,keepaspectratio,clip,trim=27 27 27 27]{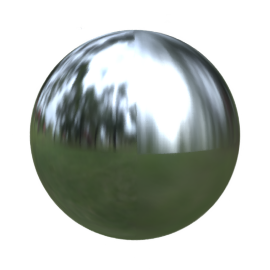} &
  \includegraphics[width=0.22\columnwidth,keepaspectratio,clip,trim=27 27 27 27]{figures/qualitative2/pred_t.png} &
  \includegraphics[width=0.22\columnwidth,keepaspectratio,clip,trim=27 27 27 27]{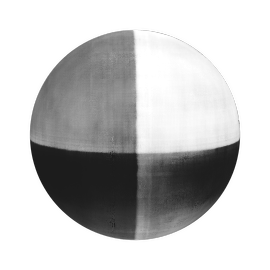} &
  \includegraphics[width=0.22\columnwidth,keepaspectratio,clip,trim=27 27 27 27]{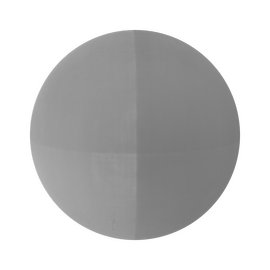} \\ \hline \\[-0.7em]

  \raisebox{3\height}{\rotatebox[origin=c]{90}{\scriptsize GT}} &
  \includegraphics[width=0.22\columnwidth,keepaspectratio,clip,trim=75 102 95 65]{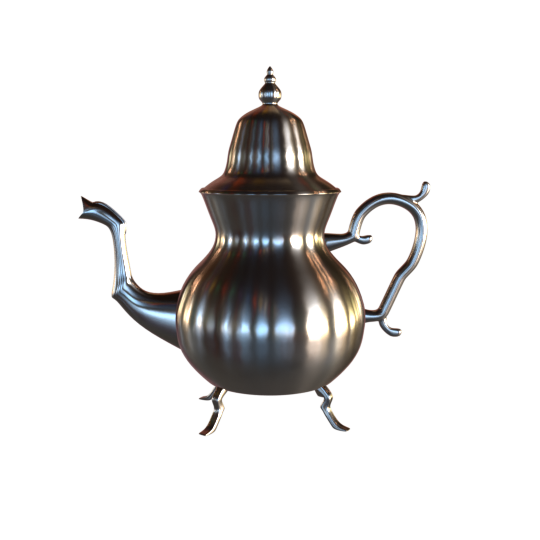}  &
  \includegraphics[width=0.22\columnwidth,keepaspectratio,clip,trim=75 102 95 65]{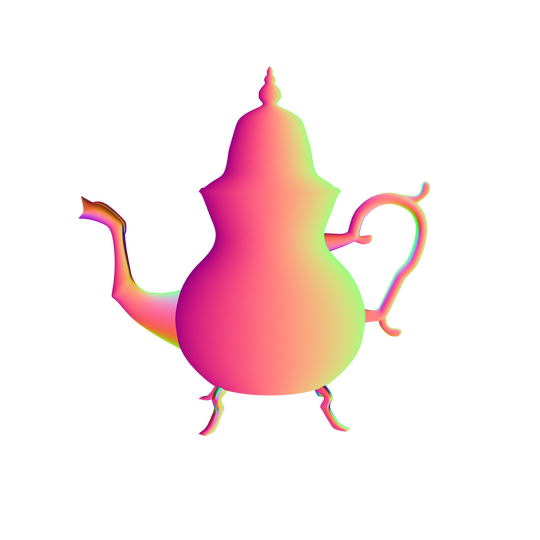}  &
  \includegraphics[width=0.22\columnwidth,keepaspectratio,clip,trim=75 102 95 65]{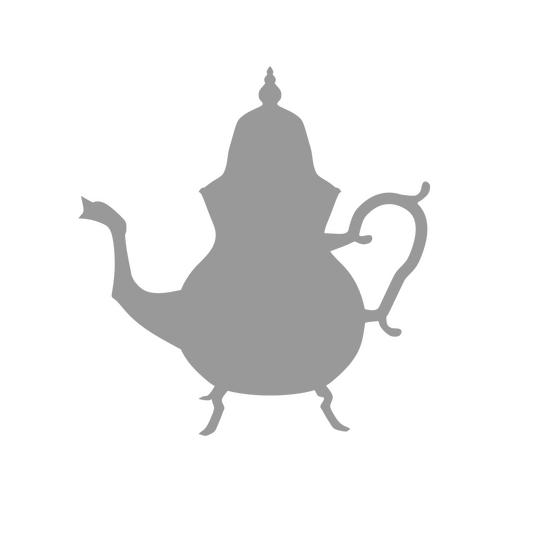}  &
  \includegraphics[width=0.22\columnwidth,keepaspectratio,clip,trim=75 102 95 65]{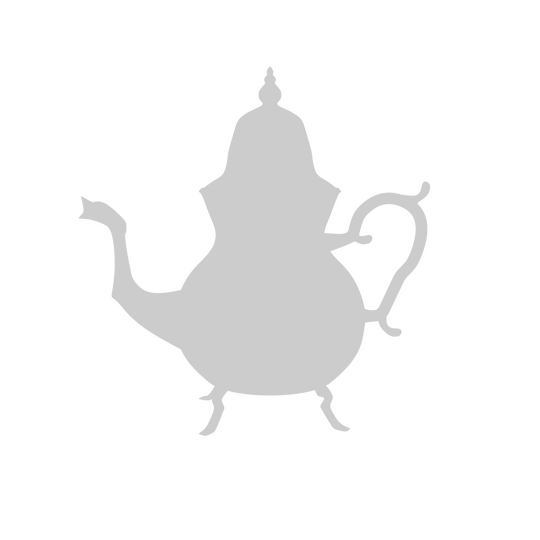}  \\
  \raisebox{1.5\height}{\rotatebox[origin=c]{90}{\scriptsize Predicted}} &
  \includegraphics[width=0.22\columnwidth,keepaspectratio,clip,trim=75 102 95 65]{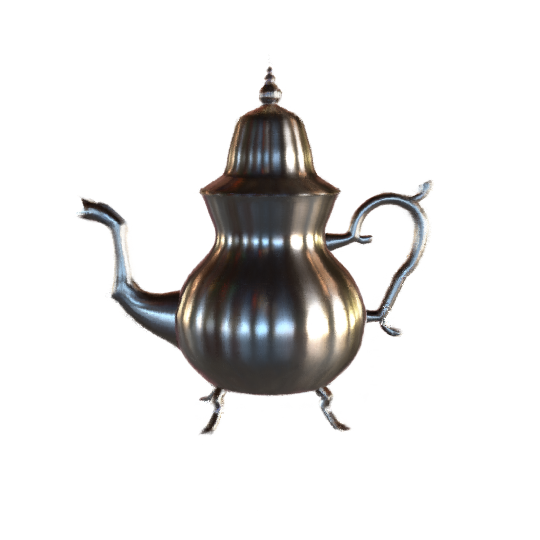} &
  \includegraphics[width=0.22\columnwidth,keepaspectratio,clip,trim=75 102 95 65]{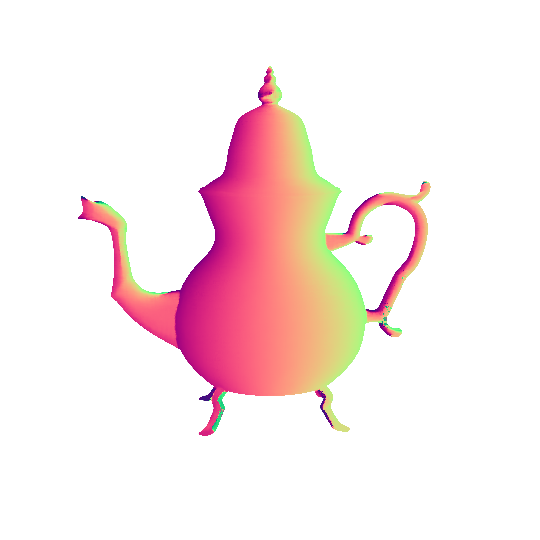} &
  \includegraphics[width=0.22\columnwidth,keepaspectratio,clip,trim=75 102 95 65]{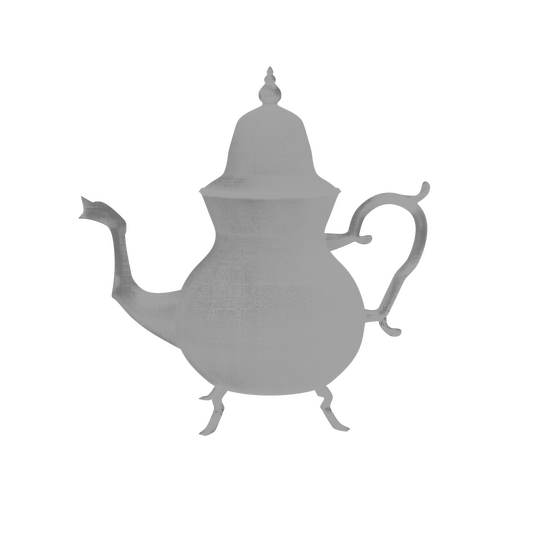} &
  \includegraphics[width=0.22\columnwidth,keepaspectratio,clip,trim=75 102 95 65]{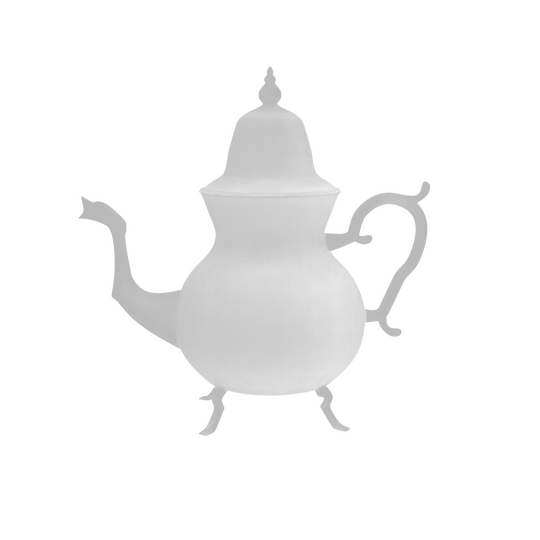} \\

\end{tabular}
  \caption{Anisotropic material properties predicted with ShinyNeRF on the ASPH spheres and the \textit{barad teapot} (CHAO). Higher intensity values correspond to larger values in $e$ and $\kappa$.}
  \label{fig:sphere_physics}
\end{figure}

The qualitative results in Figure~\ref{fig:teapot_comparison} further illustrate ShinyNeRF’s balanced performance in terms of rendering quality and geometric accuracy. While individual methods may perform best on specific scenes or metrics, ShinyNeRF is consistently able to capture sharp anisotropic highlights without noticeably compromising the reconstructed geometry. 

While all methods in Table~\ref{tab:quantitative_results} estimate some notion of specular concentration (e.g. inverse magnitude or material roughness in Ref-NeRF, sharpness parameters in AniSDF and Spec-Gaussian), only ShinyNeRF explicitly predicts surface tangents $\hat{\mathbf{t}}$ and the anisotropy coefficient $e$ at each 3D point, providing an additional level of material interpretability. As a result, it is the only approach that supports editing not only the strength and concentration, but also the direction of anisotropic specular reflections (see Figure~\ref{fig:teaser}). Note that when the material exhibits weak or no anisotropy (e.g., $e \approx 0$ or in absence of specular highlights), the ShinyNeRF is effectively unconstrained in its prediction of the tangent orientation, since any $\hat{\mathbf{t}}$ leads to the same reflectance. In such regions, the tangent MAE$^\circ$ can increase without affecting the rendered appearance. This explains why the $\hat{\mathbf{t}}$ MAE$^\circ$ on ASPH (highly specular spheres, shown in Figure~\ref{fig:sphere_physics}) is much lower compared to ASD and CHAO, where the objects exhibit weaker or more limited specular behavior.

\subsection{Limitations}

The proposed method inherits several limitations from both analytic reflectance modeling and NeRF-based optimization.

\noindent\textit{Reflectance Ambiguity} --
Recovering unique anisotropy variables from a limited number of observations is an ill-posed problem. Different anisotropy configurations can yield visually similar view-dependent reflections, leading to eccentricity and concentration estimates ($e$, $\kappa$) that deviate from the ground-truth patterns while still explaining the measured appearance. In the ASPH dataset, each sphere’s surface is divided into four sections with different ($e$, $\kappa$) combinations and only the environment map varies across spheres. As shown in Figure~\ref{fig:sphere_physics}, the predicted ($e$, $\kappa$) follow the ground-truth sections with varying fidelity, with higher alignment when the environment map provides rich texture and contrasted specular highlights are visible, which ShinyNeRF can exploit as cues for anisotropy estimation.

\noindent\textit{Bounded Anisotropy Range} --
The ASG2vMF mapping relies on a fixed, finite set of $N$ vMF lobes, which bounds the mixture capacity and limits the range of representable anisotropy. For extremely sharp or highly elongated highlights, the pre-trained network may therefore fail to reproduce the appropriate lobe shape, leading to oversmoothed estimates.

\noindent\textit{Computational Cost} --
Runtime efficiency was not a primary objective in this work. ShinyNeRF is computationally heavier than Gaussian Splatting~\cite{kerbl3Dgaussians} and other accelerated neural rendering approaches~\cite{chen2022tensorf}. Incorporating speed-oriented components such as multi-resolution hash-grid encodings~\cite{muller2022instant} could reduce training and rendering time.

\noindent\textit{Loss Weight Sensitivity} --
The hyperparameter weights in Equation~\eqref{eq:loss_complete} can substantially affect both training stability and output quality, a limitation shared with other neural rendering methods that rely on multi-term loss functions, such as Ref-NeRF and AniSDF. Adaptive loss balancing strategies, such as uncertainty-based weighting~\cite{kendall2018multi} or gradient normalization~\cite{chen2018gradnorm} could mitigate this sensitivity.
\section{Conclusion}
\label{sec:conclusion}

This work introduced ShinyNeRF, a neural radiance field framework that models both isotropic and anisotropic specular reflections by coupling volumetric rendering with an ASG-based reflectance parameterization. ShinyNeRF estimates anisotropic material properties via a pre-trained ASG2vMF network that encodes the outgoing specular radiance, enabling the rendering of sharp anisotropic highlights while recovering plausible geometry. Quantitative and qualitative evaluations on synthetic datasets with varying geometry and anisotropic complexity show that ShinyNeRF delivers competitive or state-of-the-art performance in both RGB rendering quality and geometric accuracy. Moreover, its interpretable decomposition of material properties supports direct editing of the strength, concentration, and directionality of specular reflections. These advances constitute a step towards a more faithful digital preservation of anisotropic materials, with direct applicability to tasks such as the realistic reconstruction and dissemination of cultural heritage objects.

\section*{Acknowledgements}
\label{sec:acknowledgement}

Albert Barreiro is a fellow of Eurecat's "Vicente López" PhD grant program. This work has been partially supported by the project PID2021-122136OB-C22 funded by MCIN/AEI/ 10.13039/501100011033.

{
	\begin{spacing}{1.17}
		\normalsize
		\bibliography{main_fullpaper.bbl} 
	\end{spacing}
}

\end{document}